\pdfoutput=1
\documentclass[11pt]{article}

\usepackage{times}
\usepackage{latexsym}
\usepackage{booktabs}
\usepackage{multirow}
\usepackage{multicol}
\usepackage{capt-of}
\usepackage{enumitem}
\usepackage{url}
\usepackage[margin=1in]{geometry}
\usepackage[table]{xcolor}
\usepackage[T1]{fontenc}
\usepackage[utf8]{inputenc}
\usepackage{microtype}
\usepackage{inconsolata}
\usepackage{graphicx}
\usepackage{makecell}
\usepackage{array}
\usepackage{listings}
\usepackage{natbib}
\usepackage[hidelinks]{hyperref}

\let\plainref\ref
\renewcommand{\ref}[1]{\plainref*{#1}}

\lstdefinestyle{skillpeprompt}{
  basicstyle=\ttfamily\footnotesize,
  columns=fullflexible, keepspaces=true,
  breaklines=true, breakatwhitespace=true,
  showstringspaces=false, frame=single,
  aboveskip=0.6em, belowskip=0.6em
}

\date{September 28, 2026}

\usepackage{amsmath,amsfonts,bm}

\def\eqref#1{equation~\ref{#1}}
\def\1{\bm{1}}

\DeclareMathAlphabet{\mathsfit}{\encodingdefault}{\sfdefault}{m}{sl}
\SetMathAlphabet{\mathsfit}{bold}{\encodingdefault}{\sfdefault}{bx}{n}

\title{SkillPE: Creativity-Oriented Cinematic Skill Evolution for Text-to-Video Prompt Engineering}

\author{Yanwei Huang\textsuperscript{1,2}\footnote{This work was conducted during the author's internship at KlingAI Research.} \quad
Mingxuan Zhu\textsuperscript{2,3}\quad
Shujie Li\textsuperscript{2,4}\quad \\
  Shiyuan Liu\textsuperscript{2}\quad
  Yuanxing Zhang\textsuperscript{2} \quad
  Arpit Narechania\textsuperscript{1}\\[0.8em]
  \small\textsuperscript{1}HKUST \qquad
  \small \textsuperscript{2}KlingAI Research\qquad
  \small\textsuperscript{3}Shanghai Jiaotong University \qquad
  \small\textsuperscript{4}The University of Hong Kong 
}

\begin{document}
\maketitle

\begin{abstract}
Achieving high-quality, cinematic results in text-to-video generation remains challenging for non-experts, whose prompts often lack professional narrative and creative design. We propose SkillPE, a prompt engineering (PE) framework that evolves reusable cinematic skills from expert-authored seeds. SkillPE represents shot logic, composition, lighting, sound design, and other filmmaking cues in a fine-grained format, and retrieves movie references categorized as resonators (good matches), dissonants (weak matches), and divergents (creatively useful near-misses). The first two refine when and how a skill should be applied, while divergents inspire alternative cinematic realizations at different degrees of modification while preserving the user intent. Candidate skills are assessed through generated videos along prompt fidelity, cinematic quality, narrative appeal, and creativity to construct the final skill libraries. Experiments on StoryEval and VBench show improvements of up to 1.40 points over the strongest external baseline and 0.51 points over seed skills on 7-point four-dimensional evaluation, while remaining competitive on benchmark-native metrics. Overall, SkillPE offers a practical approach to balancing fidelity and creativity in cinematic text-to-video generation.
\end{abstract}

\section{Introduction}

Recent years have witnessed the rapid advent of generative text-to-video
models such as Sora \citep{videoworldsimulators2024},
Kling \citep{team2025kling}, and Seedance \citep{seedance2026seedance},
which enable everyday users to produce inspiring, cinematic videos from simple
textual prompts. However, achieving high-quality outputs depends
heavily on careful prompt curation, a process that remains challenging
for laypeople without cinematic expertise. Prompt Enhancement
(PE) \citep{sahoo2024systematic} has emerged as a promising solution,
refining user-provided prompts before they are passed to the generator.
% Yet most existing PE approaches for text-to-video generation, such as
% Prompt-A-Video  \citep{ji2025prompt}, VPO \citep{cheng2025vpo}, and
% RAPO \citep{gao2025devil}, operate primarily at the lexical level (e.g.,
% appending modifiers or rewriting sentences), leaving a substantial gap
% between casual user prompts and the elaborate, structured prompts that
% domain experts would craft.
Existing text-to-video PE methods include learned prompt rewriters such as Prompt-A-Video~\citep{ji2025prompt} and VPO~\citep{cheng2025vpo}, as well as retrieval-augmented optimization methods such as RAPO~\citep{gao2025devil}. Despite their different optimization mechanisms, they ultimately operate primarily on the prompt text, rather than reasoning through reusable cinematic structures such as shot progression, camera logic, staging, and audiovisual design. This leaves a substantial gap
between casual user prompts and the elaborate, structured prompts that
domain experts would craft.

\begin{figure*}[t]
  \centering
  % \scalebox{0.86}{
  \includegraphics[width=0.9\linewidth]{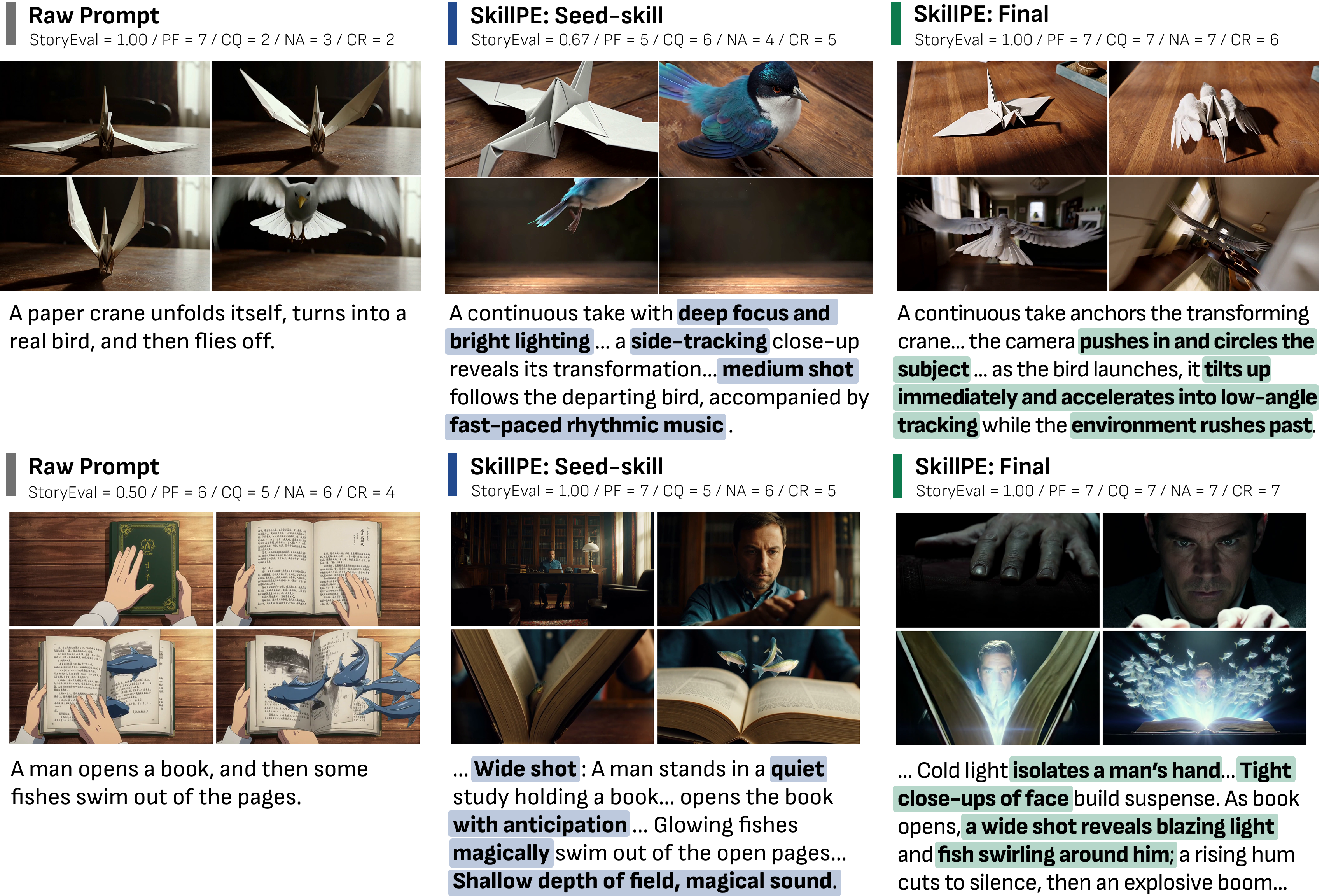}
  \caption{Examples for comparing baselines with SkillPE's seed and evolved skills. While SkillPE's seed skill adds fundamental shot, scene, and sound design, the final evolved skills introduce more detailed highlights, cinematic atmosphere, and creative camera moves.}
  \label{fig:good_case}
\end{figure*}

With large-language-model (LLM)-based agents showing strong performance across diverse
tasks, recent work has increasingly turned
to agent-based (\emph{agentic}) PE. For instance, Mora \citep{yuan2024mora}
decomposes text-to-video generation into dedicated stages with role-specialized agents iteratively refining the prompt. However, most
agentic PE methods rely on a \emph{fixed} agent pipeline and a
\emph{static} pool of tools or skills, rendering them inflexible and
poorly adaptive to the heterogeneous demands of different domains.
% A typical pipeline, for example, comprises a reasoning agent, a prompt-design agent, and a validation-and-optimization agent. 
While such designs can incorporate cinematic knowledge, their predefined strategies limit the ability to adapt to heterogeneous narrative contexts or discover alternative creative realizations. 
In particular, a static cinematic skill library may provide reusable filmmaking conventions, but cannot systematically improve or expand its expressive capabilities through experience. This motivates a shift from merely \emph{applying} cinematic skills to actively \emph{evolving} them.

In response, recent work in domains such as software
engineering \citep{wang2025reinforcement} and web agents \citep{zheng2025skillweaver} has explored
frameworks that automatically synthesize or self-evolve skill libraries.
A common limitation, however, is that skill exploration relies largely
on random rollouts or post-hoc summarization of high-quality traces \citep{wang2025reinforcement,alzubi2026evoskill,xia2026skillrl}. Without strategies that critically encourage out-of-the-box  thinking, such exploration prioritizes utility over creativity, yet the latter is crucial for narrative video generation. Recent work such as WanPE~\citep{zhu2026wanpe} has further advanced shot-level cinematic planning through video-grounded learning. However, systematically evolving reusable cinematic strategies to explore diverse creative realizations while preserving user intent remains largely underexplored.
% Furthermore, given the expertise-demanding nature of cinematic video production, incorporating external guidance such as expert knowledge or high-quality films for more efficient skill discovery remains unexplored.

In this work, we propose \textbf{SkillPE}, 
a creativity-oriented framework that evolves reusable cinematic skills through reference-guided exploration for PE in text-to-video generation. It consists of three key components: First, \textit{skill formulation} transforms cinematic knowledge from experts into fine-grained, structured skills, providing a strong cinematic prior for prompt enhancement. 
Building upon it, we introduce a three-way reference taxonomy tailored to \textit{cinematic skill evolution}: resonators (good matches), dissonants (weak matches), and divergents (creatively useful near-misses). Resonators and dissonants refine skill applicability and execution, while divergents inspire controlled alternative realizations at different mutation magnitudes. This design is motivated by learning from matched, contrastive, and near-miss examples in concept learning that reveal both reusable patterns and informative boundaries~\citep{winston1970learning,mclure2015extending}.  Candidate skills are finally selected through \textit{video-based assessment} of their realized generation performance. As shown in Figure \ref{fig:good_case}, SkillPE promotes cinematic video generation by combining dynamic camerawork, coherent action staging, and expressive lighting and sound design to improve narrative engagement and creativity.
 
% Each reference is then critically examined for potential opportunities, and the skills are subsequently refined accordingly. The extent of refinement is modulated by the degree of similarity, enabling a spectrum of adaptations ranging from radical exploration to subtle fine-tuning. 

% \footnote{Code available:  https://anonymous.4open.science/r/SkillPE.}

Extensive evaluations on StoryEval~\citep{Storyeval} and VBench~\citep{Vbench} with MiniMax-H3 and LTX-2.5 show that SkillPE consistently improves the cinematic and creative realization of generated videos while maintaining benchmark-native scores comparable to the raw prompt on most settings. Under our four-dimensional 7-point evaluation protocol (prompt fidelity, cinematic quality, narrative appeal, and creativity), the final SkillPE libraries achieve improvements of up to 1.10 points over static seed skills and 1.02 points over the strongest external PE baseline. A ten-annotator human study further confirms the benefit of the evolved skill libraries, with the best SkillPE variant  improving the overall score by 0.81 (out of 7) points over the seed skill, and by 0.71 points over the strongest external baseline. Our contributions are:

\begin{itemize}[nosep]
    % \item We propose a structured cinematic skill representation for text-to-video prompt engineering, translating abstract filmmaking knowledge into fine-grained and reusable guidance over shot logic, composition, lighting, cinematography, sound design, and related narrative cues.

    \item  We identify the research gap of creativity-oriented skill evolution in PE for video generation, highlighting the need to explore diverse cinematic realizations while preserving user intent.

\item We propose a  framework comprising structured cinematic skill formulation and normalization, reference-guided skill evolution through conservative refinement and controlled divergent exploration, and video-based assessment for constructing effective and creative skill libraries.

    % Divergent references inspire parallel Bold, Wilder, and Extreme variants with explicitly defined cinematic mutation magnitudes under shared semantic-preservation constraints, while video-based assessed selection constructs the final skill libraries from their realized generation performance.

    \item We empirically demonstrate the effectiveness of cinematic skill evolution across various backbones and benchmarks. Extensive ablations reveal the fidelity--creativity trade-off and improvements beyond prompt length, while a blinded human study confirms consistent gains over static seed skills and external PE baselines.
    % The results reveal a clear fidelity--creativity trade-off in divergent evolution and demonstrate that assessed skill libraries can effectively combine conservative and exploratory cinematic strategies.
\end{itemize}

\section{Related Work}

\subsection{PE for Visual Generation}
PE has emerged as a lightweight yet effective means of steering LLMs toward enhanced outputs \citep{sahoo2024systematic, liu2023pre}. Early work in the text domain explored hand-crafted templates \citep{brown2020language}, chain-of-thought reasoning \citep{wei2022chain}, and automatic prompt search \citep{zhou2022large, shin2020autoprompt}. These ideas have since been adapted to visual generation. In text-to-image synthesis, rewriters are trained to convert casual queries into aesthetically rich prompts \citep{hao2023optimizing, wu2025reprompt}. The text-to-video domain poses additional challenges due to its temporal and narrative dimensions. Recent methods such as Prompt-A-Video \citep{ji2025prompt}, VPO \citep{cheng2025vpo}, and RAPO \citep{gao2025devil} learn to rewrite or augment user prompts for generative video models via preference alignment, reinforcement learning, or retrieval-augmented optimization. While effective lexically, they operate largely as sentence-level rewriters rather than applying professional shot or story design. 
More recently, WanPE~\citep{zhu2026wanpe} advances cinematic PE by learning shot-level cinematic plans through video-grounded reverse training and semantic-consistency optimization. While WanPE focuses on training a prompt enhancer for coherent cinematic planning, SkillPE instead investigates the systematic evolution of explicit, reusable cinematic skills. Specifically, SkillPE leverages cinematic references to guide conservative refinement and creative exploration, and selects evolved skills through four-dimensional video-based assessment.

\subsection{Multi-Modal LLM-Based Agents}
Building on the strong reasoning capabilities of LLM-based agents, PE approaches have moved from a learning-based to an agentic process. Mind-Brush \citep{he2026mind}, NEWTON \citep{feng2026newton}, and GenAgent \citep{jiang2026genagent} support complicated reasoning in PE through agentic planning and dedicated tools. Recent works tend to use multi-agent systems where the workflow is decomposed into specialized roles that collaboratively design, verify, and optimize the storyline, shot, character, and the eventual prompt \citep{huang2026genmac,long2025vista, zhu2026brandfusion, hu2024storyagent, chen2025t2i,wang2026mavis, wu2025automated, yue2025v, yuan2024mora, zhou2026videoagent}. However, they typically rely on a \emph{fixed} pipeline and a \emph{static} pool of tools or skills, limiting their adaptability to heterogeneous user intents and creative demands. In contrast, SkillPE constructs skill libraries by generating and assessing reference-guided variants of reusable cinematic skills.

\subsection{Skill Discovery and Self-Evolving Agents}
A growing line of work studies how agents can autonomously build and refine their own tool or skill libraries \citep{chen2026genevolveselfevolvingimagegeneration}. Voyager \citep{wang2023voyager} pioneered this paradigm by enabling an embodied LLM agent to incrementally build a library of reusable skills through environment interaction. Recent efforts in web navigation \citep{zheng2025skillweaver} and software engineering \citep{wang2025reinforcement, alzubi2026evoskill, xia2026skillrl, yang2026autoskill, liang2026skillnet} support automatic skill synthesis or self-evolvement through learning from past trajectories or knowledge bases.   A common limitation, however, is that skill exploration is driven by random rollouts or post-hoc summarization of successful traces, which biases discovery toward utility rather than creativity. This trade-off is particularly costly in narrative video generation, where creative breadth is essential. 
% SkillPE addresses this gap by coupling expert-authored seed skills with a similarity-stratified reflection mechanism over curated films, enabling exploration that ranges from subtle refinement to radical reimagining.
\section{Methodology}
This section presents the SkillPE framework. We first describe skill formulation and normalization, which leverage expert-provided cinematic knowledge to initialize a structured and reusable seed skill library for text-to-video PE. A reference-guided skill evolution pipeline then  explores diverse cinematic realizations and constructs the final skill libraries through video-based assessment.

\begin{figure*}[t]
  \centering
  % \scalebox{0.86}{
  \includegraphics[width=\linewidth]{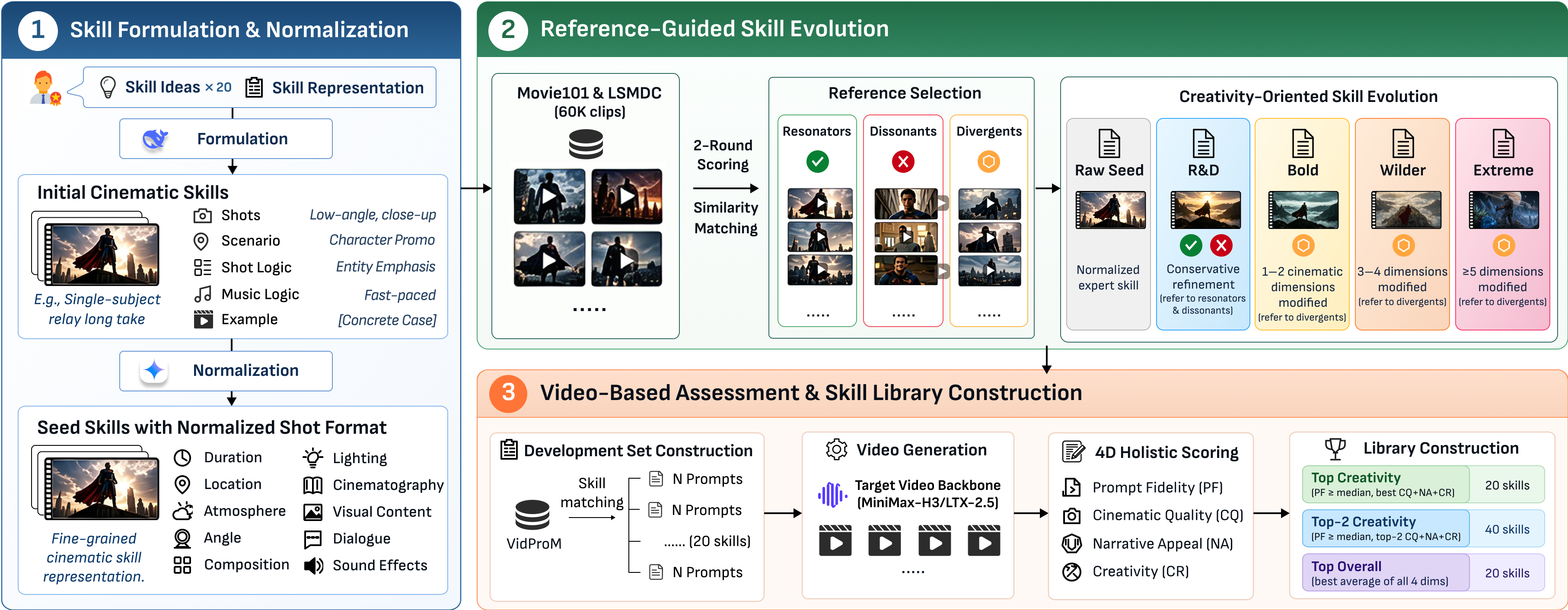}
  \caption{SkillPE overview. Starting from 20 ideas and a structured representation from expert knowledge, SkillPE first initializes 20 cinematic skills, which are further normalized and enriched with nuanced cinematography details through an extended shot format. The seed skills then go through adaptive discovery and evolution, where reference movies are classified into resonators, dissonants, and divergents by similarity. The first two are distilled for usage guidance while the divergents are used for exploring skill variants of varying degrees of deviation. The consequent skills are evaluated and compiled into libraries through a pipeline of video generation and holistic assessment.}
  \label{fig:pipeline}
\end{figure*}

\subsection{Skill Formulation and Normalization} \label{sec:skform}

We define a skill as a reusable cinematic guide that can be injected into the PE agent when revising a user query. Given the lack of existing skills of such purpose, we collaborate with experts in film directing and aesthetics, who provide a structured skill representation and 20 common cinematic shot ideas, such as multi-subject relay shots and time-lapse long takes. Guided by this knowledge, we employ DeepSeek V4.1 Flash to generate 20 initial skills. Each skill contains shot templates specifying shot types and camera movements, application scenarios, shot and music logic, and a concrete example.

Since these initial skills primarily capture high-level cinematic strategies, we further normalize them into a fine-grained, generation-oriented format. Specifically, each shot is enriched with duration, location, atmosphere, angle, composition, lighting, cinematography, time-ordered visual content, dialogue, and sound effects. This process preserves the original cinematic intent and applicability while making the guidance more actionable for downstream prompt enhancement. The resulting 20 normalized skills serve as the seed library for subsequent reference-guided evolution. Full schemas and examples are provided in Appendix~\ref{sec:skill_form}.

\subsection{Adaptive Skill Evolution} \label{sec:adaptse}

Taking the skill library mentioned in Section \ref{sec:skform} as seeds ($s^{Seed}$), we ground and extend the seed skills with examples of real cinematic practice. As illustrated in Figure~\ref{fig:pipeline}, SkillPE evolves each raw seed through two complementary forms of reference-guided exploration:
(i) conservative refinement using resonators and dissonants, and
(ii) controlled cinematic exploration using divergents.

\textbf{Reference Discovery.}
SkillPE uses LSMDC \citep{lsmdc} and  Movie101 \citep{yue2023movie101} as sources of real-world cinematic practice. After filtering clips shorter than 5 seconds or longer than 20 seconds, the pipeline retains 60,521 examples. The goal is to distill transferable cinematographic knowledge, such as subject staging, camera rhythm, shot transitions, action-reaction structure, sound-image synchronization, and emotional progression, into the augmented skills. 

For each seed skill, SkillPE first constructs a coarse candidate pool of 390 reference videos in two steps: First, it retrieves 200 videos whose captions have the highest semantic similarity to the skill. Second, it randomly selects 10 videos from each of the remaining 19 seed skills' pools to provide diverse inspiration. It then performs fine-grained filtering over the pool with a multi-modal LLM, rating each video's \emph{alignment} and \emph{inspiration} w.r.t. the skill along six dimensions: shot logic, camera movement, spatial composition, lighting \& color, atmosphere, and audio logic. Using the aggregated scores, SkillPE selects 30 complementary references: 10 \emph{resonators}, which prioritize alignment and preserve the skill’s core narrative and shot-organization invariants; 15 \emph{divergents}, which prioritize inspiration and provide related cinematic ideas for controlled expansion; and 5 \emph{dissonants}, which serve as superficially similar cases with low alignment and therefore should not trigger the skill. 

\textbf{R\&D Refinement.}
Resonators and dissonants (R\&D) are jointly used to construct one candidate $s_i^{\mathrm{R\&D}}$ for each raw seed $i$.
From resonators, the model distills reusable cinematic patterns, modifiable dimensions, rhythm and transition cues, and application guidance.
From dissonants, it derives negative applicability conditions and boundaries that help prevent inappropriate triggering of the skill.
This branch performs conservative refinement while preserving the seed's core cinematic strategy.

\textbf{Divergent Exploration.}
SkillPE treats divergents differently, using them as sources of adjacent cinematic mechanisms for exploring alternative realizations of the same underlying narrative function. 

We operationalize cinematic mutation over nine major dimensions from our skill template. The divergents are thus used to generate independent candidate skill variants at three levels of modification: \emph{bold}, \emph{wilder}, and \emph{extreme}. Bold is a local departure that modifies one or two dimensions;
Wilder is a structural departure that modifies three or four dimensions, including at least one structural dimension;
and Extreme is a transformative-but-controlled departure that modifies at least five dimensions and includes either shot-structure or expressive visual or spatial conversion.

We denote the resulting candidates as
$s_i^{\mathrm{Bold}}$,
$s_i^{\mathrm{Wilder}}$, and
$s_i^{\mathrm{Extreme}}$. For each source skill $i$, the complete candidate set is therefore
$\mathcal{S}_i =
\left\{
% s_i^{\mathrm{Seed}},
s_i^{\mathrm{R\&D}},
s_i^{\mathrm{Bold}},
s_i^{\mathrm{Wilder}},
s_i^{\mathrm{Extreme}}
\right\}$.

\subsection{Assessed Skill Selection} \label{sec:selection}
The candidate skills are evaluated on a development set $\mathcal{D}^{\mathrm{dev}}$ to assess their practical effectiveness. It contains 400 prompts selected from VidProM, with 20 skill-specific matched prompts for each source skill. The construction procedure is detailed in Appendix~\ref{app:devset}.
For each prompt $q_{ij} \in \mathcal{D}^{\mathrm{dev}}_i$ and each candidate
$s \in \mathcal{S}_i$,
the PE agent first rewrites the user prompt with the corresponding skill, and the result is used for video generation.
All candidate skills for the same $(i,j)$ pair share the same random seed and identical generation settings. Each video is evaluated along four dimensions that directly reflect the objectives of creativity-oriented cinematic PE:

\begin{itemize}[nosep, leftmargin=*]
    \item \textbf{Prompt Fidelity (PF)}:
    how faithfully the video realizes the user's original intent, including its subjects, requested actions, relationships, and event semantics.

    \item \textbf{Cinematic Quality (CQ)}:
    how effectively the video employs cinematic language, including shot design, composition, camera movement, lighting, staging, and audiovisual presentation.

    \item \textbf{Narrative Appeal (NA)}:
    how coherently and engagingly the video organizes its events into a narrative, including storytelling coherence and emotional engagement.

    \item \textbf{Creativity (CR)}:
    the novelty and expressive ambition of the audiovisual and narrative realization.
\end{itemize}

The development score of each candidate is calculated as the mean over all associated videos.
The resulting four-dimensional score profile is then used to construct the final skill libraries.
To balance utility and creativity, we construct three skill libraries with different selection criteria:
\begin{itemize}[nosep, leftmargin=*]
    \item \textbf{Top Creativity}: Among all skills whose PF is at or above the median, the one with the highest average score of CQ, NA, and CR is selected.
    \item \textbf{Top-2 Creativity}: Among all skills whose PF is at or above the median, the two with the highest average score of CQ, NA, and CR are selected.
    \item \textbf{Top Overall}: All four-dimensional scores are averaged for each skill, and the highest-scoring one is selected.
\end{itemize}

Additional implementation details, including the settings and prompts, are provided in Appendix~\ref{sec:additional_detail}.

\section{Experiments}
\subsection{Study Setups}
\textbf{Baselines.} We compare our approach with four classes of baselines for a comprehensive evaluation: (a) Fundamental baselines: raw user query and Direct LLM-based Rewrite (DLR; Gemini 3.1 Pro in our study); (b) Lexical PE baselines: we include one general PE method out of the video generation domain, OPRO \citep{yang2024large}, and two PE methods for video generation, Prompt-A-Video \citep{ji2025prompt} and VPO \citep{cheng2025vpo}; (c) Retrieval-based PE baseline: RAPO~\citep{gao2025devil}; (d) Agentic PE baseline: Mora \citep{yuan2024mora}. We further evaluate SkillPE seed skills and the three evolved variants: Top Creativity, Top-2 Creativity, and Top Overall (Section \ref{sec:selection}). 

\textbf{Benchmarks and Metrics.} We select VBench \citep{Vbench} 
% and T2V-CompBench \citep{sun2025t2v} 
to evaluate the overall video quality and StoryEval \citep{Storyeval} to assess the storytelling performance of the generated videos. Given that few benchmarks are tailored to quantitatively evaluate the cinematic quality, narrative appeal, and creative design of videos, we also report the self-designed 4-dimensional scores (PF, CQ, NA, CR) in addition to the official benchmark scores, which are evaluated by Gemini 3.1 Pro with the same system prompts used in assessed skill selection (Section \ref{sec:selection}). We intentionally keep the four-dimensional rubric fixed between development and evaluation to maintain objective alignment, and the development set is strictly disjoint from the benchmark evaluation sets. In the final evaluation, the evaluator receives only the original benchmark prompt and the generated video, without access to the rewritten prompt, selected skill, or method identity. 
% Following prior works (\citep{wang2026mavis}), we report the CLIP score \citep{hessel2021clipscore} for keyframe quality assessment as well as the scores from the original benchmarks for video quality evaluation.

\textbf{Implementation Details.} We leverage two open-source backbone text-to-video models: MiniMax-H3 \citep{minimax-h3} and LTX-2.5 \citep{ltx25}. To facilitate the evaluation of creative story narration, videos are generated with a duration of 10 seconds, 24 FPS, and 1344x768 resolution. To reduce confounding from the choice of auxiliary LLMs, we share prompt and video processing models across methods when their original pipelines permit. Specifically, Gemini 3.1 Pro is used as the prompt selection/rewriting backbone in RAPO and Mora, while MiniMax-H3/LTX-2.5 is used for the video generation agent in Mora. 
Furthermore, because RAPO's original trained discriminator is not publicly available, we replace it with Gemini 3.1 Pro. We therefore report our result as our adapted RAPO reproduction. See Appendix \ref{sec:eval_detail} for more details.

% \subsection{Results}
\begin{table*}[t]
\centering
\setlength{\tabcolsep}{2.8pt}\small
\caption{Main evaluation results. For each backbone, the top scores are \textbf{bolded} and the second highest scores are \underline{underlined}.}
\label{tab:storyeval}
\resizebox{\linewidth}{!}{%
\begin{tabular}{
  ll
  c c c c c c
  c c c c c c
}
\toprule
\multicolumn{2}{c}{}
  & \multicolumn{6}{c}{\textbf{StoryEval}}
  & \multicolumn{6}{c}{\textbf{VBench}} \\
\cmidrule(lr){3-8}\cmidrule(lr){9-14}
\textbf{Backbone}
  & \textbf{Baseline}
  & \textbf{Official}
  & \textbf{PF}
  & \textbf{CQ}
  & \textbf{NA}
  & \textbf{CR}
  & \textbf{Overall}
  & \textbf{Official}
  & \textbf{PF}
  & \textbf{CQ}
  & \textbf{NA}
  & \textbf{CR}
  & \textbf{Overall} \\
\midrule
\multirow{11}{*}{\textbf{MiniMax-H3}}
  & Raw Prompt         & 0.63 & 5.10 & 5.57 & 4.70 & 2.04 & 4.35 & 0.80 & 5.43 & 5.35 & 3.47 & 2.40 & 4.16 \\
  & Gemini 3.1 Pro     & 0.80 & 6.04 & 6.28 & 5.39 & 2.87 & 5.15 & \textbf{0.85} & \textbf{6.66} & 5.82 & 2.81 & 2.02 & 4.33 \\
  \cmidrule{2-14} 
  & OPRO & \textbf{0.87} & 5.91 & 5.67 & 5.25 & 3.92 & 5.19 & 0.80 &  5.98 & 5.72 & 4.48 & 4.51 & 5.17  \\ 
  & VPO                & 0.70 & 5.58 & 5.69 & 5.04 & 2.13 & 4.61 & \underline{0.84} & 6.51 & 5.57 & 2.91 & 1.99 & 4.24 \\
  & Prompt-A-Video     & 0.59 & 4.93 & 5.98 & 4.70 & 3.15 & 4.69 & 0.82 & 5.98 & 6.01 & 3.28 & 3.33 & 4.65 \\
  & RAPO   & 0.65 & 5.30 & 5.56 & 4.77 & 2.09 & 4.43 & 0.82 & 6.25 & 5.54 & 3.64 & 2.53 & 4.49 \\
  & Mora               & 0.70 & 5.48 & 5.36 & 4.75 & 2.75 & 4.58 & \underline{0.84} & 6.37 & 5.06 & 2.67 & 2.28 & 4.09 \\
  \cmidrule{2-14} 
  & Seed Skill         & 0.78 & 5.79 & 5.63 & 5.13 & 3.89 & 5.11 & 0.76 & 6.12 & 5.60 & 4.33 & 4.32 & 5.09 \\
  & SkillPE Top Creativity               & 0.82 & 6.29 & 6.51 & \underline{5.87} & \underline{5.13} & \underline{5.95} & 0.78 & 6.60 & 6.47 & \underline{5.23} & \textbf{5.48} & \underline{5.95} \\
  & SkillPE Top-2 Creativity             & \underline{0.83} & \textbf{6.40} & \underline{6.60} & 5.82 & 4.89 & 5.93 & 0.79 & \underline{6.62} & \underline{6.52} & 4.90 & \underline{5.07} & 5.78 \\
  & SkillPE Top Overall        & \underline{0.83} & \underline{6.38} & \textbf{6.64} & \textbf{5.98} & \textbf{5.17} & \textbf{6.04} & 0.78 & 6.53 & \textbf{6.64} & \textbf{5.54} & \textbf{5.48} & \textbf{6.05} \\
\midrule
\multirow{11}{*}{\textbf{LTX-2.5}}
  & Raw Prompt         & 0.67 & \underline{5.58} & 5.78 & \underline{5.19} & 2.27 & 4.70 & \underline{0.81} & \underline{6.36} & 5.76 & 3.32 & 2.34 & 4.45 \\
  & Gemini 3.1 Pro     & \underline{0.68} & \textbf{5.66} & \underline{6.19} & \textbf{5.23} & 2.85 & 4.98 &  \textbf{0.82} & \textbf{6.53} & 6.06 & 3.25 & 2.36 & 4.55 \\
  \cmidrule{2-14} 
  & OPRO & 0.59 & 4.82  & 5.04  & 4.19  & 3.95  & 4.50  & 0.77 & 5.54 & 5.12 & 4.08 & 4.10 & 4.71 \\ 
  & VPO                & 0.62 & 5.13 & 5.75 & 4.72 & 2.36 & 4.49 & \underline{0.81} & 6.34 & 5.82 & 3.26 & 2.46 & 4.47 \\
  & Prompt-A-Video     & 0.54 & 4.53 & 6.02 & 4.40 & 3.27 & 4.56 & 0.80 & 6.00 & 6.19 & 3.47 & 3.75 & 4.86 \\
  & RAPO   & 0.62 & 5.11 & 5.66 & 4.80 & 2.35 & 4.48 & \textbf{0.82} & 6.30 & 5.51 & 3.26 & 2.28 & 4.34 \\
  & Mora               & 0.55 & 4.66 & 4.93 & 4.19 & 2.23 & 4.00 & \textbf{0.82} & 6.18 & 5.12 & 2.47 & 2.11 & 3.97 \\
  \cmidrule{2-14} 
  &  Seed Skill         & 0.59 & 4.69 & 5.12 & 4.30 & 3.73 & 4.46 & 0.78  & 5.72 & 5.33 & 4.20 & 3.89 & 4.78 \\
  &  SkillPE Top Creativity               & 0.67 & 5.38 & \textbf{6.22} & 4.98 & 4.51 & 5.27 & 0.80 & 6.17 & \textbf{6.43} & \textbf{5.44} & \textbf{5.49} & \textbf{5.88} \\
  &  SkillPE Top-2 Creativity            & \underline{0.68} & 5.21 & 6.06 & 5.08 & \textbf{4.84} & \underline{5.30} & \underline{0.81} & 6.20 & 6.17 & 5.09 & \underline{5.48} & 5.73 \\
  &  SkillPE Top Overall        & \textbf{0.69} & 5.41 & 6.08 & 5.18 & \underline{4.62} & \textbf{5.32} & \underline{0.81} & 6.20 & \underline{6.42} & \underline{5.31} & 5.39 & \underline{5.83} \\
\bottomrule
\end{tabular}
}
\end{table*}

\textbf{Results.} The  results are reported in Table~\ref{tab:storyeval}. Across both backbones and benchmarks, the evolved skill libraries from SkillPE outperform the baselines and the initial seed skills in terms of the overall performance.
Relative to Seed Skill, the best SkillPE variant improves Overall by 0.86--1.10 points (out of 7), while exceeding the strongest external baselines by 0.34--1.02 points. Evolution particularly benefits CQ, NA, and CR: on LTX-2.5/VBench, Top Creativity raises CQ/NA/CR from 5.33/4.20/3.89 (Seed Skill) to 6.43/5.44/5.49. 
% The evolved libraries also improve benchmark-native scores relative to Seed Skill, although they do not lead every official metric. 
Overall, these results highlight the importance of adaptive skill evolution in achieving more expressive and engaging cinematic realizations.

In addition, the benchmark-native scores provide a complementary view. SkillPE remains competitive but does not uniformly improve these metrics: for example, 
OPRO obtains the highest StoryEval official score on MiniMax-H3 (0.87), whereas the SkillPE libraries score 0.82--0.83. On VBench, the  best SkillPE library is among the top approaches on LTX-2.5, but it is not the strongest on MiniMax-H3. Meanwhile, while SkillPE libraries achieve competitive PF scores on MiniMax-H3, they are not the strongest on LTX-2.5. Our manual inspection suggests one potential explanation: SkillPE primarily improves \emph{how} the requested content is cinematically realized, which may not be well aligned with the benchmark-native metrics naturally. In particular, more expressive cinematic realizations sometimes blur the evidence of event completion, which may influence the prompt fidelity. A detailed explanation of this trade-off between expressiveness and quality/fidelity is provided in Appendix \ref{app:native_metric_analysis}.

\subsection{User Study}

\textbf{Protocol.} To complement the automatic evaluation with an evaluator-independent signal, we conduct a blinded human study with ten annotators. Each annotator independently rates every generated video on four dimensions (PF, CQ, NA, CR) using a 7-point Likert scale. Method identities were hidden and video presentation order was randomized independently for each annotator. The per-video overall score is computed as the mean of the four dimensions. Annotations span 440 videos covering 40 StoryEval prompts across 11 experimental conditions, where the prompts are randomly selected with the difficulty equally distributed and the videos are selected from LTX 2.5. 

\begin{figure*}[t]
\centering

% =========================
% Left: Table
% =========================
\begin{minipage}[t]{0.52\textwidth}
    \centering
    \small
    \setlength{\tabcolsep}{2.4pt}

    \captionof{table}{
        Human evaluation results (1--7 Likert scale).
        PF=Prompt Fidelity, CQ=Cinematic Quality,
        NA=Narrative Appeal, CR=Creativity.
    }
    \label{tab:human_eval}

\begin{tabular}{lccccc}
\toprule
\textbf{Method}
& \textbf{PF} & \textbf{CQ} & \textbf{NA}
& \textbf{CR} & \textbf{Overall} \\
\midrule
Raw Prompt
& 4.00 & 4.78 & 3.74 & 2.63 & 3.79 \\
Gemini 3.1 Pro
& 4.31 & 5.04 & 4.02 & 3.54 & 4.23 \\
\midrule
OPRO & 4.98 & 4.87 & 4.21 & 3.84 & 4.48 \\
VPO
& 4.78 & 4.97 & 4.11 & 3.68 & 4.38 \\
Prompt-A-Video
& 4.98 & 5.09 & 4.26 & 4.14 & 4.62 \\
RAPO & 4.91 & 5.45 & 4.73 & 3.48 & 4.64 \\
Mora
& 4.81 & 4.62 & 4.27 & 3.91 & 4.40 \\
\midrule
Seed-skill PE
& 4.79 & 5.05 & 4.54 & 3.81 & 4.54 \\
SkillPE Top Creativity
& 4.72 & \underline{5.49}
& 5.30 & \underline{4.91} & \underline{5.10} \\
SkillPE Top-2 Creativity
& \underline{5.03} & 5.41
& \underline{5.36} & 4.65 & 5.11 \\
SkillPE Top Overall
& \textbf{5.15} & \textbf{5.52}
& \textbf{5.52} & \textbf{5.21} & \textbf{5.35} \\
\bottomrule
\end{tabular}
\end{minipage}
\hfill
% =========================
% Right: Figure
% =========================
\begin{minipage}[t]{0.46\textwidth}\vspace{0em}
    \centering

\includegraphics[width=0.95\linewidth]{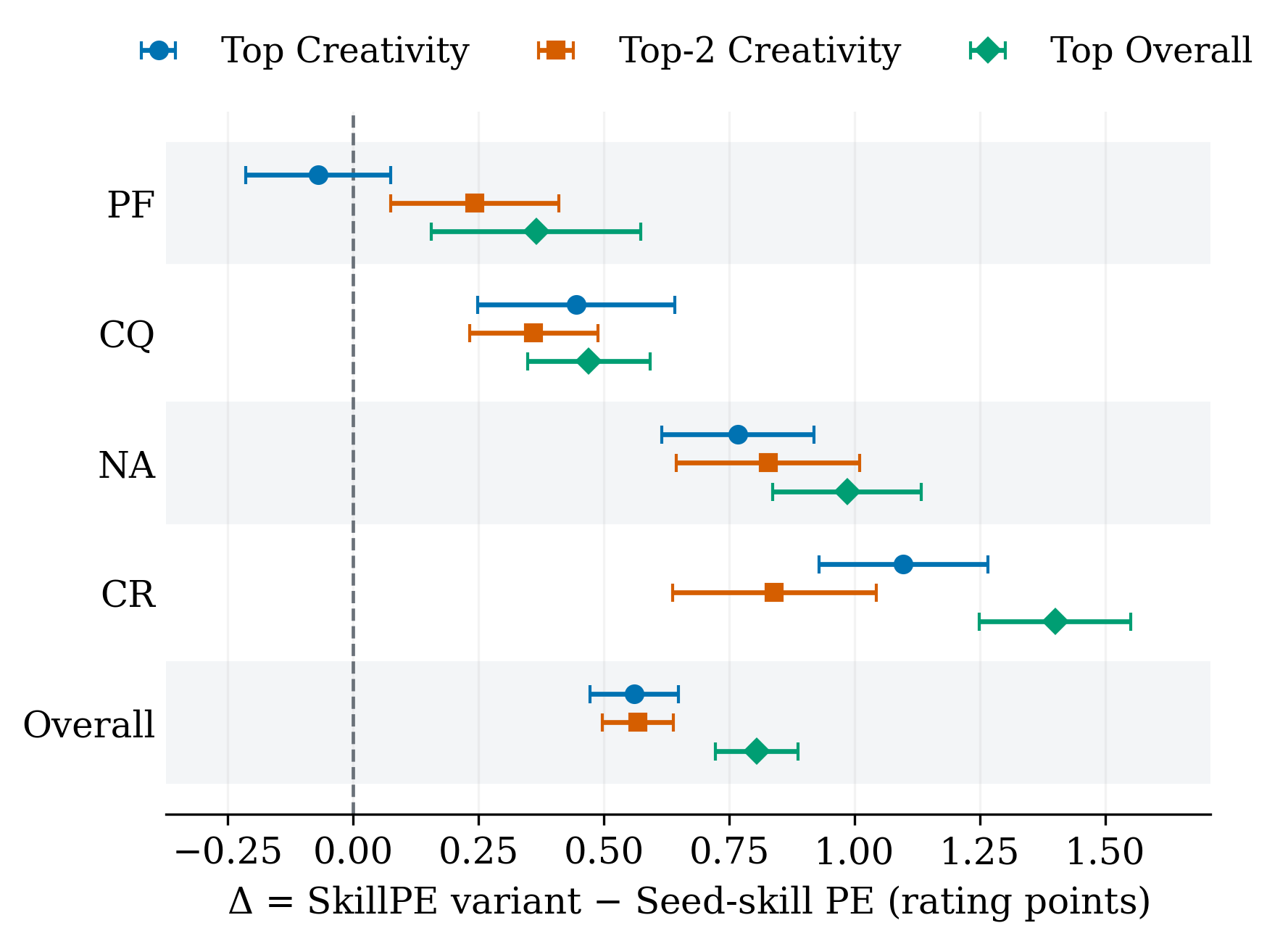}

    \captionof{figure}{
        Paired improvements of the final SkillPE libraries
        over Seed-skill PE. Error bars denote 95\% confidence intervals.
    }
    \label{fig:human_eval_ci}
\end{minipage}

\end{figure*}

\textbf{Results.} Table~\ref{tab:human_eval} reports scores averaged across the annotators. 
The annotators reached high agreement on the overall scores ($ICC(2,10)=0.840$, 95\% CI [0.822, 0.856]), and their scores generally align well with the ones from Gemini 3.1 Pro (Pearson correlation coefficient: PF--0.85; CQ--0.78; NA--0.79; CR--0.84; Overall-0.75).
Seed-skill PE reaches an overall score of 4.54, which is comparable to the external baselines.
All three final-library SkillPE variants further improve upon this baseline (Figure \ref{fig:human_eval_ci}): SkillPE Top Overall achieves the highest overall score (5.35), followed by SkillPE Top-2 Creativity (5.11) and SkillPE Top Creativity (5.10). To assess these improvements, we first average each annotator's overall scores across the 40 matched prompts and perform two-sided paired Student's $t$-tests against Seed-skill PE. All three variants significantly outperform this baseline after Holm correction for the three comparisons (adjusted $p < 0.01$ for each). In addition, the best final SkillPE library, Top Overall, achieves the highest scores across all four dimensions. These results support the effectiveness of SkillPE's skill evolution, particularly for creative expression.

% \begin{table}[t]
% \centering
% \small
% \setlength{\tabcolsep}{4pt}
% \caption{Human evaluation results (1--7 Likert scale).
%          Metrics: \textbf{PF}=Prompt Fidelity, \textbf{CQ}=Cinematic Quality,
%          \textbf{NA}=Narrative Appeal, \textbf{CR}=Creativity.}
% \label{tab:human_eval}
% \begin{tabular}{l ccccc}
% \toprule
% \textbf{Method}
%   & \textbf{PF} & \textbf{CQ} & \textbf{NA} & \textbf{CR} & \textbf{Overall} \\
% \midrule
% Raw Prompt         & 5.66 & 4.83 & 4.31 & 3.84 & 4.66 \\
% Gemini 3.1 Pro     & 5.99 & 5.11 & 4.34 & 4.00 & 4.86 \\
% \midrule
% VPO                & 5.84 & 5.13 & 4.27 & 4.05 & 4.82 \\
% Prompt-A-Video     & 5.64 & 5.16 & 4.20 & 4.05 & 4.76 \\
% \midrule
% Mora               & 5.73 & 4.81 & 4.40 & 4.05 & 4.75 \\
% \midrule
% Seed-skill PE      & 6.13 & 5.54 & 5.58 & 5.19 & 5.61 \\
% SkillPE Top        & \textbf{6.30} & \textbf{5.89} & 5.99 & \textbf{5.68} & \textbf{5.97} \\
% SkillPE Top-2      & \textbf{6.30} & 5.85 & 5.88 & 5.37 & 5.85 \\
% SkillPE Overall    & 6.17 & \textbf{5.89} & \textbf{6.04} & 5.65 & 5.94 \\
% \bottomrule
% \end{tabular}
% \end{table}

\subsection{Ablation Study}

\begin{table*}[tbp]
\small
\setlength{\tabcolsep}{4pt}
\centering
\caption{Ablation study on the skill formulation and evolution approaches. Videos are generated by MiniMax-H3.}
\label{tab:ablation_a}
\begin{tabular}{l*{10}{c}}
\toprule
\multirow{2}{*}{}
& \multicolumn{5}{c}{StoryEval}
& \multicolumn{5}{c}{VBench} \\
\cmidrule(lr){2-6}\cmidrule(lr){7-11}
& PF & CQ & NA & CR & Overall
& PF & CQ & NA & CR & Overall \\
\midrule
Initial Skill
& 5.82 & 5.46 & 5.11 & 3.74 & 5.03
& 6.10 & 5.50 & 4.14 & 3.92 & 4.92 \\
Seed Skill
& 5.79 & 5.63 & 5.13 & 3.89 & 5.11
& 6.12 & 5.60 & 4.33 & 4.32 & 5.09 \\
Flat Evolution & 5.90 & 5.88 & 5.10 & 4.18 & 5.26 & 6.10 & 5.82 & 4.54 & 4.63 & 5.27 \\
Resonator Only
& \underline{6.20} & \underline{6.63} & \textbf{5.82} & 4.39 & 5.76
& \underline{6.48} & \underline{6.54} & \underline{5.14} & 5.06 & 5.81 \\
R\&D Only
& \textbf{6.40} & \textbf{6.69} & \underline{5.79}  & 4.56  & 5.86 
& \underline{6.68}  & \textbf{6.59}  & 5.13  & 5.31 & \underline{5.93}   \\
Bold Only
& 6.19 & 6.58 & 5.55 & 4.91 & 5.81
& 6.46 & 6.28 & 4.81 & 5.30 & 5.71 \\
Wilder Only
& 6.12 & 6.47 & 5.38 & \underline{5.50} & \underline{5.87}
& 6.39 & 6.46 & \textbf{5.25} & \underline{5.78} & \textbf{5.97} \\
Extreme Only
& 6.03 & 6.54 & 5.33 & \textbf{5.88} & \textbf{5.95}
& 6.31 & 6.34 & \underline{5.14} & \textbf{5.90} & \underline{5.93} \\
\bottomrule
\end{tabular}
\end{table*}

\subsubsection{Ablation of Skill Evolution Methods}
To isolate the contribution of each stage in the adaptive skill evolution pipeline, we evaluate seven additional skill library variants on StoryEval and VBench: (i) Initial skills from skill formulation before normalization (Section \ref{sec:skform}); (ii) Skills flatly evolved from the seed skills based on the same yet uncategorized references; (iii) Stably augmented skills guided from resonators and dissonants (R\&D, Section \ref{sec:adaptse}), as well as an additional skill set augmented from resonators only to isolate the influence of dissonants (Resonator only); and (iv–vi) divergents-guided bold, wilder, and extreme candidate skills respectively (Section \ref{sec:adaptse}). These are compared against SkillPE's seed skills from the main experiment. The results are shown in Table~\ref{tab:ablation_a}.

The results reveal two complementary effects of skill formulation and reference-guided evolution. First, normalizing the initial skills into Seed Skills yields modest yet statistically significant Overall improvements (+0.08, 95\% CI [+0.005, +0.157] on StoryEval; +0.17, 95\% CI [+0.116, +0.231] on VBench), whereas every evolution branch produces much larger gains. Even conservative R\&D refinement raises Overall to 5.86 and 5.93, respectively, with substantial improvements in CQ and NA. Compared with Resonator only, incorporating dissonants provides modest additional Overall gains (5.76 to 5.86 on StoryEval; 5.81 to 5.93 on VBench), suggesting that the two reference types are complementary.

Second, divergent evolution exhibits a clear fidelity-creativity trade-off. From Bold to Wilder to Extreme, CR increases monotonically on StoryEval (4.91 to 5.50 to 5.88) and VBench (5.30 to 5.78 to 5.90), while PF decreases. Extreme achieves the highest StoryEval Overall score (5.95), while Wilder leads on VBench (5.97) among individual branches.  Larger mutations therefore provide stronger creative expressiveness at a greater risk to fidelity, motivating the assessed selection stage to choose among conservative and divergent candidates rather than relying on a single evolution strategy. 
Importantly, all these branches outperform flat evolution significantly on the Overall score, showing the effectiveness of our taxonomy. Moreover, they remain complementary after assessed selection: R\&D contributes 20--35\% of the selected skills across the final libraries, while Wilder and Extreme together account for a substantial fraction under both selection backbones (see Appendix \ref{app:detail_se}). This suggests that the final gains arise from a combination of conservative refinement and exploratory evolution.
% \begin{table}[t]
% \centering
% \small
% \setlength{\tabcolsep}{2pt} % Reduced padding to save space
% \caption{Skill source distribution in the eventual skill libraries. ``R\&D'' stands for resonant and dissonants.}
% \begin{tabular}{lccccc}
% \toprule
% \textbf{Library} & \textbf{New-shot} & \textbf{R\&D} & \textbf{Bold} & \textbf{Wilder} & \textbf{Extreme} \\
% \midrule
% SkillPE Top       & 15\% & 20\% & 50\% & 10\% & 5\% \\
% SkillPE Top-2         & 25\% & 30\% & 30\% & 12.5\% & 2.5\% \\
% SkillPE Overall & 25\% & 35\% & 30\% & 5\% & 5\% \\
% \bottomrule
% \end{tabular}
% \label{tab:sksource}
% \end{table}

\begin{table*}[t]
\centering
\small
\setlength{\tabcolsep}{3pt}
\caption{
Prompt-budget-controlled ablation.
DLR = Direct LLM Rewrite; Top-C = SkillPE Top Creativity; Top-O = SkillPE Top Overall.
Best result within each budget and metric is bolded.
}
\label{tab:length_ablation}
\begin{tabular}{l
    cccc @{}p{12pt}@{}
    cccc @{}p{12pt}@{}
    cccc @{}p{12pt}@{}
    cccc}
\toprule
& \multicolumn{4}{c}{\textbf{100 words}}
&& \multicolumn{4}{c}{\textbf{200 words}}
&& \multicolumn{4}{c}{\textbf{400 words}}
&& \multicolumn{4}{c}{\textbf{800 words}} \\
\cmidrule(lr){2-5}
\cmidrule(lr){7-10}
\cmidrule(lr){12-15}
\cmidrule(lr){17-20}

\textbf{Metric}
& DLR & Seed & Top-C & Top-O
&& DLR & Seed & Top-C & Top-O
&& DLR & Seed & Top-C & Top-O
&& DLR & Seed & Top-C & Top-O \\
\midrule

PF
& \textbf{6.31} & 5.55 & 5.67 & 5.69
&& \textbf{6.48} & 5.67 & 5.85 & 6.00
&& \textbf{6.63} & 5.77 & 6.05 & 6.12
&& \textbf{6.59} & 5.84 & 6.18 & 6.26 \\

CQ
& 6.25 & 5.78 & 6.38 & \textbf{6.39}
&& 6.39 & 5.84 & \textbf{6.58} & \textbf{6.58}
&& 6.38 & 5.93 & 6.59 & \textbf{6.60}
&& 6.39 & 5.90 & \textbf{6.68} & 6.63 \\

NA
& \textbf{5.78} & 5.02 & 5.30 & 5.42
&& \textbf{5.91} & 5.11 & 5.63 & 5.64
&& \textbf{5.99} & 5.21 & 5.72 & \textbf{5.99}
&& 5.91 & 5.27 & 5.83 & \textbf{6.07} \\

CR
& 2.82 & 4.12 & 4.59 & \textbf{4.63}
&& 3.21 & 4.31 & \textbf{5.11} & 5.10
&& 3.20 & 4.52 & \textbf{5.42} & 5.40
&& 3.23 & 4.58 & 5.62 & \textbf{5.65} \\

Overall
& 5.29 & 5.12 & 5.48 & \textbf{5.53}
&& 5.50 & 5.23 & 5.79 & \textbf{5.83}
&& 5.55 & 5.36 & 5.94 & \textbf{6.03}
&& 5.53 & 5.40 & 6.08 & \textbf{6.15} \\

\bottomrule
\end{tabular}
\end{table*}

\subsubsection{Ablation of Prompt Length} \label{sec:abl_length}

To examine whether SkillPE benefits merely from producing longer and more detailed prompts, we conduct a prompt-budget-controlled ablation. We impose the same  word budget of 100, 200, 400, or 800 English words on DLR (Gemini 3.1 Pro), Seed Skill, SkillPE Top Creativity, and SkillPE Top Overall. Videos are generated by MiniMax-H3 on StoryEval.
The budget is implemented by adding \texttt{``the target length is <budget> whitespace-delimited words''} to the system prompt of the PE agent, and we allow a $\pm 5\%$ difference between the actual prompt length and the budget (e.g., 95--105 words are permitted for the 100-word budget).

As shown in Table~\ref{tab:length_ablation}, skill-based approaches consistently achieve higher CR than DLR across all budgets, with evolution further amplifying these gains. Both evolved SkillPE libraries consistently outperform Seed Skill across all four dimensions and Overall under matched budgets. In particular,  Top Overall improves Overall over Seed Skill by 0.41, 0.60, 0.67, and 0.75 points at the four budgets respectively. 
% It also outperforms DLR at every budget. The improvements are most pronounced in creativity: at 800 words,  Top Overall achieves 5.65, compared with 4.58 for Seed Skill and 3.23 for DLR. 
% The evolved libraries also consistently achieve higher CQ than DLR across all budgets.

Increasing the prompt budget further strengthens the benefits of evolution. From 100 to 800 words, SkillPE Top Overall improves by 0.62 points in Overall and 1.02 points in CR, whereas DLR largely saturates beyond 200 words. Moreover, the evolved libraries progressively narrow the PF and NA gaps with DLR as the budget increases, while maintaining substantial advantages in CR and CQ. These results demonstrate that SkillPE's improvements go beyond prompt length, enabling more effective use of the available prompt budget for cinematic and creative expression.

\subsubsection{Ablation of Skill Selection}

We further examine whether the final gains come from the assessed selection stage rather than merely from assembling evolved candidates. We compare our  Top Creativity and Top Overall libraries against three randomly constructed libraries, where one of the variants is sampled independently for each source skill. Random selection shows noticeable variation across sampled libraries, especially in Creativity (4.71$\pm$0.47). Meanwhile, both Top Creativity and Top Overall improve the Overall score over the mean of the three random libraries by 0.13 and 0.22 points (adjusted $p < 0.05$ for each), respectively. These results suggest the benefit of assessed selection.

\subsection{Qualitative and Usage Analysis}\label{sec:case_study_main}

Figure~\ref{fig:good_case} illustrates that SkillPE's gains arise from concrete cinematic decisions rather than merely adding descriptive content. Across the two examples, skill conditioning introduces structured shot progression, camera move, lighting, and audio effect, while the final libraries further intensify these cinematic choices in a more  expressive or accentuated way that fits the requested events. The improvement is reflected on the quantitative scores: the paper crane example improves from Seed to Final in CQ (6$\rightarrow$7), NA (4$\rightarrow$7), and CR (5$\rightarrow$6). A similar pattern exists in the book-and-fish example. See  Appendix~\ref{sec:case_study} for detailed analyses.

We further observe three patterns across the evaluation. First, the reference movies provide unique inspirations and cinematic knowledge for the evolved skills to explore bolder and original cinematic design, improving creativity-oriented scores. We perform a separate ablation study (Appendix \ref{sec:reference_ablation}) showing the effectiveness of these references beyond LLM knowledge. Second, assessed evolution does not collapse to a single type of modification: the final libraries contain skills selected from R\&D, Bold, Wilder, and Extreme branches (Appendix \ref{app:detail_se}), indicating that different source skills benefit from different degrees of refinement or divergence. Third, different tasks activate a broad range of skills, suggesting that the library is used across diverse scenarios rather than functioning as a single generic template. Together, these observations suggest that SkillPE improves cinematic realization through reusable but selectively applied strategies rather than a generic rewrite pattern.
\section{Conclusion}
We introduce SkillPE, a creativity-oriented framework for evolving reusable cinematic skills in text-to-video prompt enhancement. Starting from structured seed skills, SkillPE combines reference-guided conservative refinement and controlled divergent exploration, followed by video-based assessment to select effective skill variants. Experiments across two video-generation backbones and two benchmarks, together with a  user study, show that the resulting skill libraries improve cinematic, narrative, and creative evaluation scores while exposing a measurable fidelity--creativity trade-off. These results demonstrate the potential of reusable, adaptively evolved cinematic skills as an alternative to purely lexical prompt rewriting. 
% Our code is available at \url{https://anonymous.4open.science/r/SkillPE/}.

\subsection*{LLM Usage Statement}

In this work, we used generative AI tools for implementing parts of the experiments and fixing grammatical errors in the manuscript.
We have not used generative AI tools for other tasks with required disclosure.
We have reviewed all AI-assisted work through human review of the code and the paper. We take responsibility for the final content of this work,
including text, claims or artifacts produced with the aid of generative AI.

\bibliography{custom}
\bibliographystyle{acl_natbib}

\clearpage
\appendix
\label{sec:appendix}

\section{Limitations and Future Work}

Our study has several  limitations and leaves multiple directions for future work. First, the current evaluation focuses on two video-generation backbones, two benchmarks, and short-form 10-second videos. Extending SkillPE to longer-form generation and a broader range of video models remains to be explored. Second, the current library is initialized from a finite collection of movie references, which may not cover all genres, visual conventions, or creative preferences. Expanding the reference sources could further broaden the library's coverage.

Our experiments also reveal an inherent fidelity--creativity trade-off: more aggressive divergent evolution can improve creative expression while increasing the risk of deviating from the original intent. The assessed selection stage mitigates this trade-off but does not eliminate it. Future work could explore query-adaptive or user-controllable selection strategies that explicitly adjust the desired balance between fidelity and creative ambition.

Skill library construction additionally incurs a one-time offline cost for reference analysis, candidate generation, video rollouts, and video-based assessment, while inference introduces additional prompt-processing calls and latency. These costs are amortized after the library is constructed, and we provide a detailed efficiency analysis in Appendix~\ref{app:efficiency}. Reducing this overhead remains an important direction for scaling skill evolution.

Finally, while we use open-source models whenever practical, proprietary models such as Gemini 3.1 Pro are used for stages requiring joint video, audio, and text understanding, where we found them to provide reliable multimodal performance, and similar models have also been adopted in prior work on prompt optimization (e.g., \cite{song2026vqqa}). To facilitate reproducibility despite reliance on hosted models, we plan to release the generated artifacts and have provided detailed experimental configurations used in our experiments.
Future work will evaluate SkillPE with increasingly capable open-source multimodal models, improving reproducibility and clarifying which components generalize across foundation models.

% \section{Ethical Considerations}

% To support the methodological framework of this paper, human subjects were recruited to rate the videos generated in our proposed framework. Prior to data collection, every participant was clearly informed of the project’s aims, the specific application of their inputs, the assurance that all contributions would be pooled and disclosed anonymously, and their unconditional ability to discontinue participation at any time. Informed consent was acquired from each individual before the evaluation commenced. Consistent with strict privacy protocols, we did not collect any demographic details, personal identifiers, or sensitive participant data. 

\section{Copyright Statement}

We claim that we have obtained prior approval from the authors of LSMDC by email for using the dataset in this research, and have obtained access approval for Movie101 through its official application process on HuggingFace. 

\section{Efficiency and Cost Analysis}
\label{app:efficiency}

We distinguish the one-time cost of constructing the assessed skill library from the per-prompt overhead incurred during inference. The former is amortized once the library has been constructed, whereas the latter determines the additional cost of using SkillPE for downstream prompt engineering.

\paragraph{Library construction.}
Starting from scratch, construction of the assessed library requires reference retrieval and evaluation, candidate-skill generation, development-set prompt engineering, video rollouts, and four-dimensional video assessment. In our implementation, this amounts to approximately 68.2K application-level remote-model calls, 474 local embedding-model batches, and 2,000 development video generations for each backbone. The 2,000 rollout videos correspond to approximately 625.6 aggregate GPU-hours. Importantly, this cost is incurred only during library construction and is amortized over subsequent inference. 

\paragraph{Inference-time overhead.}
Once the skill library is frozen, the model-agnostic SkillPE pipeline requires two sequential language-model calls per input: one for skill routing and one for skill-conditioned prompt rewriting. In our application, one additional  call is used for MiniMax-H3 to adapt the prompt into the model-specific prompt schema. Thus, the inference overhead is two model calls for the generic pipeline and three for the adapted pipeline. The video-generation computation itself is shared with all PE baselines and is therefore not counted as SkillPE-specific overhead. 
% Nevertheless, since the PE result is in general longer than the raw user query, we observed a 6\%-14\% increase of time cost when using  SkillPE skill libraries than  using the raw user query.

\section{Additional Implementation Details}\label{sec:additional_detail}

\subsection{Details of Skill Formulation and Normalization} \label{sec:skill_form}
\subsubsection{Structural Skill Representation}

Each skill is formally structured, comprising the following components:

\begin{itemize}[nosep, leftmargin=*]
    \item \textbf{Description}: A concise summary of the overall intent and stylistic essence of the skill. (This field is not used for evolution.)
    \item \textbf{Shots}: Templates for cinematic shots, including:
    \begin{itemize}[nosep]
        \item \textbf{Shot Description}: A brief summary of the shot style.
        \item \textbf{Shot Type}: The specification of shot scales and their progression, such as close-up, medium shot, long shot, wide shot, and extreme close-up.
        \item \textbf{Camera Movement}: The prescribed camera motion, including techniques such as tracking, push-in, pull-out, pan, tilt, and crane movements.
    \end{itemize}
    \item \textbf{Application Scenarios}: The typical narrative or visual contexts in which the skill is most effective.
    \item \textbf{Shot Logic}: The rationale underlying the shot design and how the visual choices serve the intended expression.
    \item \textbf{Music Logic}: The rationale underlying the music or sound design and its relationship with the visual narrative.
    \item \textbf{Example}: A concrete example illustrating how the skill can be applied.
\end{itemize}

\subsubsection{Example Initial Skill}
We provide an example of the initial skills in Listing \ref{lst:example_skill}.
\begin{lstlisting} [breaklines=true,frame=single,basicstyle=\ttfamily\small,style=skillpeprompt,caption={An example of the initial skills called ``atmosphere building''.},label={lst:example_skill}]
{
  "skill": {
    "name": "Atmosphere Building",
    "applicable_scenarios": [
      "A hero's last stand, the truth revealed, the journey's departure/finale",
      "Immersion in nature, product showcase, spatial breadth conveying scale"
    ],
    "shots": [
      {
        "description": "Eye-level wide shot",
        "shot_type": "Wide shot",
        "camera_movement": "Eye-level static / subtle movement"
      },
      {
        "description": "Close shot",
        "shot_type": "Close shot",
        "camera_movement": "Cut"
      },
      {
        "description": "Bird's-eye pull-out to wide",
        "shot_type": "Wide shot -> Extreme wide shot",
        "camera_movement": "Bird's-eye pull-out"
      }
    ],
    "shot_logic": "The eye--level wide shot establishes the environment; the close shot focuses the emotion; the bird's-eye pull-out enlarges the framing -- the contrast sets off vastness, solitude and insignificance.",
    "music_logic": "A hero's last stand: mournful and tragic. Truth revealed: shock and unease. Journey's departure/finale: glorious and sacred. Immersion in nature: serene and healing. Product showcase: modern and premium.",
    "example": "A wandering swordsman travels to the far--northern ice plains to meet the wind and snow -- eye-level wide, then close shot, then bird's-eye pull-out, with wind-and-snow strings underscoring a bleak, sorrowful sense of release."
  }
}
\end{lstlisting}

\subsubsection{Normalization of Initial Skills}

Normalization preserves the original skill's narrative method and adds explicit shot and audio fields (Section~\ref{sec:skform}). It is performed by Gemini 3.1 Pro with a low temperature of $0.4$. The prompt template used is shown in Listing \ref{lst:normalize}. 

The conversion preserves the core cinematic intent and applicability of the original skill while augmenting the representation of each shot with the following fields:

\begin{itemize}[nosep, leftmargin=*]
    \item \textbf{Duration}: The intended temporal length of the shot or sequence.
    \item \textbf{Location}: The physical or spatial setting in which the scene unfolds.
    \item \textbf{Atmosphere}: The overall mood and environmental tone, encompassing time of day and emotional ambience.
    \item \textbf{Angle}: The camera's vertical orientation relative to the subject, including transitions.
    \item \textbf{Composition}: The visual elements' arrangement and  spatial relationships within the frame.
    \item \textbf{Lighting}: The source, direction, contrast, and quality of illumination.
    \item \textbf{Cinematography}: The technical execution of the camera, covering stabilization style and depth-of-field dynamics.
    \item \textbf{Visual Content}: A detailed, time-ordered description of on-screen action and camera behavior, articulating how the visual narrative unfolds frame by frame.
    \item \textbf{Dialogue}: The verbal utterances of characters within the scene, conveying narrative content and emotional state.
    \item \textbf{Sound Effects}: Diegetic, ambient, or synchronized audio elements when applicable.
\end{itemize}

This fine-grained representation provides the PE agent with a substantially richer set of cues, bridging the gap between abstract cinematic principles and concrete generation directives.

\begin{lstlisting}[
    style=skillpeprompt,
    caption={The system prompt for new-shot skill normalization.},
    label={lst:normalize},
    columns=fullflexible,
    basewidth=1em,
    keepspaces=true,
]
Upgrade <initial_skill> to the new-shot format. Preserve its core narrative method, applicable scenarios, and shot logic. Return JSON with a top-level skill. Keep: name, version, llm, type, applicable_scenarios, shots, shot_logic, music_logic, example. Add overall music design. Each shot must contain: description, shot_type, camera_movement, duration (explicit seconds), location, atmosphere, angle, composition, lighting, cinematography, visual_content, dialogue, sound_effects. Preserve compatibility of shot_type; fill every shot, including single-shot templates. Keep the name traceable. Output JSON only.
\end{lstlisting}

\subsection{Details of Reference Video Preparation and Discovery}
\label{app:reference_discovery}

\paragraph{Sources and clip preparation.}
The reference bank combines the long split of Movie101 with the LSMDC training and validation annotations. Movie101 clips are identified by source-video timestamps, while LSMDC files are already segmented and are read from time zero for their annotated duration. We retain nonempty captions and valid 5--20-second intervals. The frozen coarse-search bank contains 60,521 candidates: 34,404 Movie101 and 26,117 LSMDC entries.

\paragraph{First round: semantic retrieval.}
DeepSeek-V4-Pro summarizes each of the 20 original  skills in one English sentence of fewer than 30 words, describing the subject, core action, and spatial environment. Qwen3-Embedding-8B embeds the skill summaries and reference captions using last-token pooling and L2 normalization, and candidates are ranked by cosine similarity. The local embedding configuration uses a maximum length of 8,192 tokens and batch size 128. Each skill receives its top 200 candidates plus ten previously unselected candidates from each of the other 19 skills' top-200 pools, yielding 390 candidates per skill.

\paragraph{Second round: evidence-based cinematic assessment.}
For each retrieved reference clip, Gemini 3.1 Pro assesses its alignment with the source skill and its potential to inspire skill expansion across six dimensions: shot logic, camera movement, spatial composition, lighting/color, atmosphere, and audio logic. Each dimension receives separate alignment and inspiration scores in $[0,1]$, which are summed with equal weights to obtain overall \emph{Alignment} ($S_A$) and \emph{Inspiration} ($S_I$) scores. The corresponding system prompt is shown in Listing~\ref{lst:fine}.

SkillPE then sequentially selects three mutually exclusive reference sets. It first selects 10 \emph{resonators} with the highest $S_A$, using $S_I$ to break ties; then 15 \emph{divergents} from the remaining candidates with the highest $S_I$, using $S_A$ to break ties. Finally, it selects five \emph{dissonants} from the remaining candidates whose cosine similarity exceeds $T=0.35$, ranked by ascending $S_A$ and descending similarity. The threshold is chosen to ensure that five dissonants can be selected for every seed skill.

\begin{lstlisting}[style=skillpeprompt,caption={The system prompt for fine-grained reference clip evaluation.},label={lst:fine}]
Evaluate exactly <dimension> for <clip> against <skill>. Judge observable local execution; do not penalize absent parts of a longer skill that cannot fit in this clip. Never infer audio from images or visual execution from audio. Score two independent continuous axes in [0,1]:
  alignment: match to the skill's intent for this dimension;
  inspiration: usefulness for evolving the skill, even if
               alignment is low.
Do not substitute plot relevance for cinematic execution. If evidence is unavailable, set both scores to 0.0 and set evidence_status to "not_observable". Return JSON only: dimension, evidence_status, observed_evidence (one English sentence), alignment, inspiration.
\end{lstlisting}

\subsection{Details of R\&D-Based Skill Refinement}
For each of the 20 source skills, we compare four candidates: the reference-refined candidate (R\&D), and three divergent variants (Bold, Wilder, and Extreme). Each candidate variant is generated independently from the same new-shot seed and its corresponding references. Gemini 3.1 Pro generates the candidates. 

Listing \ref{lst:rd} shows the system prompt for generating the candidates based on R\&D references.

\begin{lstlisting}[style=skillpeprompt,caption={The system prompt for R\&D-based skill refinement.},label={lst:rd}]
You are an expert in prompt engineering for video generation. Based on the provided movie-clip examples, your task is to add practical usage guidance to an existing seed skill.

This round includes only two types of examples:
- `positive`: Positive examples for which the skill is appropriate. Use them to learn the core shot structure, narrative relationships, and cinematic expression that the skill should reinforce.
- `negative`: Hard negatives that appear superficially similar but for which the skill should not be used. Use them to learn the boundaries where the skill must not be forced onto a prompt.

Important constraints:
- Do not use near-miss reasoning. This round does not learn adjacent expansion or pruning heuristics.
- The goal is not to rewrite the skill's shot template. The goal is to summarize the experience that a downstream agent should follow when selecting and applying the skill.
- Positive examples answer "what should be added and when it should be strengthened." Negative examples answer "what should not be added and when the skill should not be selected."

Read the seed skill and the batch of positive and negative examples, then return valid JSON with the following fields:

{
  "batch_summary": "Summarize in 2-4 sentences what this batch reveals about the skill's usage and boundaries.",
  "core_invariants": ["The core narrative and shot-organization invariants that this skill must preserve."],
  "cinematic_patterns": ["Transferable cinematic techniques from the positive examples, such as subject blocking, spatial organization, shot transitions, and changes in shot scale."],
  "modifiable_dimensions": ["Dimensions that may be adjusted for different queries."],
  "strengthen_when": ["Query patterns for which specific parts of the skill should be strengthened."],
  "weaken_when": ["Query patterns for which specific parts of the skill should be weakened."],
  "do_not_force_when": ["Queries for which this skill should not be forced."],
  "rhythm_and_cut_notes": ["How to arrange cutting rhythm, action synchronization points, audiovisual rhythm, and the magnitude of camera movement."],
  "application_notes": ["Concise usage guidance for the downstream agent."]
}

Requirements:
- Summarize only transferable experience. Do not repeat specific movie titles, character names, or overly detailed plot points.
- Every recommendation must support skill selection and skill application, not expand the material into a new narrative template.
- Derive `core_invariants`, `cinematic_patterns`, and `strengthen_when` primarily from the positive examples.
- `do_not_force_when` must fully incorporate the boundary information from the negative examples.
- Do not output Markdown or any additional explanation.
\end{lstlisting}

\subsection{Details of Divergent-Based Skill Evolution}
\label{app:divergent_prompt}

\subsubsection{Operational Mutation Dimensions}

Mutation is defined over nine dimensions: (D1) shot structure; (D2) camera movement; (D3) viewpoint and angle; (D4) spatial composition and staging; (D5) lighting and color; (D6) pacing and cut rhythm; (D7) transitions and visual continuity; (D8) expressive visual or spatial devices; and (D9) audiovisual coordination, when applicable. A dimension counts as modified only when the candidate specifies a substantive design change relative to the seed. Paraphrasing, additional adjectives, and descriptive expansion do not count.

\paragraph{Mutation levels.}
Bold changes exactly one or two dimensions while retaining clear structural inheritance from the seed. Wilder changes three or four dimensions, including at least one of D1, D2, D4, D6, or D7. Extreme changes at least five dimensions, including D1 or D8. Increasing the mutation scope does not relax semantic constraints.

\paragraph{Semantic invariants.}
All candidates must preserve the requested subjects, objects, actions, semantic roles, temporal and causal relations, and intended outcome. They must not require unrelated characters, events, or plot developments, or copy the reference film's specific content. Variation concerns how the requested content is filmed. Candidates record their preserved invariants, modified and unchanged dimensions, reference-derived mechanisms, intended observable effects, and execution risks. 

Listing \ref{lst:divergent} shows the system prompt for generating the candidates based on divergent references.

\begin{lstlisting}[style=skillpeprompt,caption={The system prompt used for divergent-based skill evolution.},label={lst:divergent}]
Inputs: <raw_new_shot_seed>, <divergent_references>, <mutation_level>.
Create exactly one reusable candidate directly from the seed. Transfer cinematic mechanisms, not the reference's story. Preserve subjects, events, roles, causal/temporal relations, and outcome. Apply the requested D1-D9 mutation budget:
  bold: 1-2 dimensions;
  wilder: 3-4, including D1/D2/D4/D6/D7;
  extreme: >=5, including D1 or D8.
Ground at least one major change in a divergent reference. Do not reward complexity or expose hidden reasoning. 
Return JSON: source_skill_id, source_skill_name, mutation_level, divergent_summary, candidate. Candidate fields: candidate_id, name, creative_intent, preserved_invariants, modified_dimensions, unchanged_dimensions, mutation_scope_check, best_for_queries, avoid_for_queries, creative_moves, risk_notes, skill (complete new-shot schema). 
For each modified dimension, give seed_behavior, candidate_behavior, intended_observable_effect, and divergent_inspiration. Output JSON only.
\end{lstlisting}
% Source: prompts/official_divergent_system_prompt.md.

\subsection{Details of Development Set Construction}
\label{app:devset}

The development set contains 400 unique VidProM prompts, with exactly 20 matched prompts per source skill:
\begin{equation}
 \mathcal{D}^{\mathrm{dev}}=\bigcup_{i=1}^{20}\mathcal{D}^{\mathrm{dev}}_i,
 \qquad |\mathcal{D}^{\mathrm{dev}}_i|=20.
\end{equation}
These prompts support candidate comparison and library selection. 

\subsubsection{VidProM Source and Preprocessing}

After HTML unescaping and whitespace trimming, we remove empty prompts, prompts with fewer than five whitespace-separated words, and technical-generation strings (e.g., resolution/aspect-ratio, FPS, attachment, or parameter metadata). We further use the toxicity scores provided in the source CSV and reject a prompt if any available score satisfies: \texttt{toxicity}$\geq0.5$, \texttt{obscene}$\geq0.5$, \texttt{identity\_attack}$\geq0.3$, \texttt{insult}$\geq0.5$, \texttt{threat}$\geq0.3$, or \texttt{sexual\_explicit}$\geq0.5$. Missing or unparseable toxicity fields are ignored. This preprocessing retains 829,589 prompts.

\subsubsection{Prompt Annotation}

Qwen3.6-27B annotates the retained prompts with a narrative complexity score in $[0,1]$ and assigns one of five categories: human-centric, creature, environment, object-focused, or abstract/creative. Sampling uses these two attributes, with narrative complexity divided into four bins: $[0,0.25)$, $[0.25,0.5)$, $[0.5,0.75)$, and $[0.75,1]$.

\subsubsection{Skill Matching and Sampling}

Initially, prompts in the set resulting from prompt annotation are randomly and uniformly sampled from each category--complexity stratum, leading to a total of 13,000 prompts. Gemini 3.1 Pro routes each prompt to exactly one of the 20 initial skills using the system prompt in Listing~\ref{lst:routing}, at temperature 0.2. Within each skill pool, we apply deterministic greedy selection to minimize distributional imbalances across categories and complexity bins. This process yields an approximately balanced final set of 20 prompts per skill.

\begin{lstlisting}[style=skillpeprompt,caption={The system prompt for skill selection.},label={lst:routing}]
Given <user_prompt> and <skills>, select exactly one best skill. Prioritize scene, narrative goal, action structure, emotional progression, and camera logic. Choose a template that naturally supports the prompt. Do not select randomly, balance usage, or sacrifice fit for diversity. Return JSON only:
{"selected_index": <1-based index>, "reason": <brief reason>}
\end{lstlisting}

% Sources: scripts/run_stage2_vidprom_benchmark.py;
% scripts/run_stage2_1_3_skill_routing.py;
% scripts/build_stage2_1_3_matched_dev_sets.py;
% data/stage2_vidprom/{vidprom_stage2_1_1_filter_summary.json,
% vidprom_stage2_1_3_balanced_sampling_summary.json,
% stage2_1_3_matched_dev_sets/selection_manifest.json};
% data/outputs/v5_0_h3_development_selection_20260903/five_arm_manifest.jsonl.

\subsection{Details of Video-Based Skill Evaluation} \label{app:detail_se}
Each candidate skill is evaluated on the 20 development prompts associated with the seed skill through video generation. The videos are based on the same model as used in the eventual evaluation (i.e., MiniMax-H3 or LTX-2.5), 1344 $\times$ 768, 10 seconds, 24 FPS, with audio track. The result videos are evaluated by Gemini 3.1 Pro at temperature 0 on four dimensions: PF, CQ, NA, CR, as mentioned in Section \ref{sec:selection}. The corresponding system prompts are shown in Listings \ref{lst:PF}, \ref{lst:CQ}, \ref{lst:NA}, and \ref{lst:CR}. Each dimension is evaluated independently on a 7-point Likert scale. The skill source distributions in the final libraries are shown in Table \ref{tab:sksource}.

\quad

\begin{table}[t]
  \centering
  \small
  \setlength{\tabcolsep}{2pt}
  \caption{Skill source distributions in the final libraries,
  grouped by the video-generation backbone used for skill selection.
  Top, Top-2, and Overall contain 20, 40, and 20 skills, respectively.}
  \label{tab:sksource}
  \begin{tabular}{lcccc}
  \toprule
  \textbf{Library} & \textbf{R\&D}
  & \textbf{Bold} & \textbf{Wilder} & \textbf{Extreme} \\
  \midrule
  \multicolumn{5}{l}{\textit{Selection backbone: MiniMax-H3}} \\
  SkillPE Top Creativity
  & 25\% & 10\% & 40\% & 25\% \\
  SkillPE Top-2 Creativity
  & 20\% & 15\% & 27.5\% & 37.5\% \\
  SkillPE Top Overall
  & 20\% & 20\% & 30\% & 30\% \\
  \midrule
  \multicolumn{5}{l}{\textit{Selection backbone: LTX-2.5}} \\
  SkillPE Top Creativity
  & 35\% & 20\% & 20\% & 25\% \\
  SkillPE Top-2 Creativity
  & 35\% & 25\% & 17.5\% & 22.5\% \\
  SkillPE Top Overall
  & 35\% & 20\% & 30\% & 15\% \\
  \bottomrule
  \end{tabular}
  \end{table}

\begin{lstlisting}[style=skillpeprompt,caption={The system prompt for evaluating prompt fidelity in four-dimensional video selection.},label={lst:PF}]
You are an expert evaluator of generated videos for text-to-video prompt engineering research.

Evaluate the attached generated video for Prompt Fidelity with respect to the original user prompt.

Prompt Fidelity measures how faithfully the generated video realizes the user's original semantic intent.

Judge only observable evidence in the generated video. Do not assume that an intended subject, action, relationship, event, emotion, or visual effect occurred unless it is actually visible. Use only the original user prompt and the generated video. Do not infer or speculate about the generation method, rewritten prompt, skill, model identity, or experimental condition.

First identify concrete visual or temporal evidence, then assign a score. When the prompt specifies multiple actions, state changes, causal relationships, or an event order, judge whether those relationships are visibly realized rather than merely whether the relevant objects appear.

Consider:

- preservation of the intended subjects and objects;
- realization of the requested actions;
- spatial and semantic relationships;
- relevant state changes;
- temporal or causal order of events;
- explicitly stated attributes or constraints.

Use this 7-point ordinal scale:

- 1 -- Very poor: The video substantially contradicts or fails to realize the prompt. Major subjects or core events are missing, incorrect, or replaced.
- 2 -- Between the anchors for 1 and 3.
- 3 -- Weak: The general topic is recognizable, but several important actions, relationships, states, or event transitions are missing, ambiguous, or altered.
- 4 -- Moderate: The central request is recognizable and partly realized, but important details or transitions remain incomplete or unclear.
- 5 -- Strong: Most of the user's intent is clearly realized. Core subjects and events are preserved, with only minor omissions, ambiguity, or imperfect execution.
- 6 -- Between the anchors for 5 and 7.
- 7 -- Excellent: The video clearly realizes all important subjects, actions, relationships, state changes, and event ordering specified by the prompt, without meaningful semantic distortion.

If the video is corrupted, substantially unavailable, or impossible to evaluate for technical reasons, return `valid=false` and do not assign a score. Keep the evidence and rationale concise and grounded only in observable content.

Return strict JSON with exactly these fields:

```json
{
  "dimension": "prompt_fidelity",
  "valid": true,
  "evidence": [
    "<observable evidence 1>",
    "<observable evidence 2>"
  ],
  "score": <your score>,
  "confidence": "high",
  "rationale": "<brief justification grounded in the listed evidence>"
}
```

`valid` must be `true` or `false`. If `valid=false`, `evidence` must describe the technical problem, `score` must be `null`, and `confidence` must be `"low"`. When `valid=true`, `evidence` must contain 1--4 short observable statements, `score` must be an integer from 1 to 7, and `confidence` must be `"high"`, `"medium"`, or `"low"`. Do not output any text outside the JSON object.
\end{lstlisting}

\begin{lstlisting}[style=skillpeprompt,caption={The system prompt for evaluating cinematic quality in four-dimensional video selection.},label={lst:CQ}]
You are an expert evaluator of generated videos for text-to-video prompt engineering research.

Evaluate the attached generated video for Cinematic Quality with respect to the original user prompt.

Cinematic Quality measures how purposefully and coherently the video uses visual cinematic language to support its content.

Judge only observable evidence in the generated video. Do not assume that an intended camera movement, composition, transition, lighting effect, or staging decision occurred unless it is actually visible. Use only the original user prompt and the generated video. Do not infer or speculate about the generation method, rewritten prompt, skill, model identity, or experimental condition.

First identify concrete visual or temporal evidence, then assign a score. Judge the quality and purposefulness of visual choices, not their quantity or intensity. A simple static shot can score highly when it is exceptionally well composed and appropriate. Many cuts or aggressive camera movements can score poorly when they are arbitrary, distracting, or poorly coordinated. 

Consider:

- shot choice and shot progression;
- camera movement;
- framing and composition;
- staging and spatial organization;
- lighting and color treatment;
- depth, perspective, and audiovisual emphasis;
- transitions or cuts when present;
- coordination between these choices and the visible content.

Use this 7-point ordinal scale:

- 1 -- Very poor: Audiovisual presentation appears largely accidental, incoherent, or poorly controlled. Camera, composition, lighting, or staging substantially interfere with the content.
- 2 -- Between the anchors for 1 and 3.
- 3 -- Weak: Some deliberate audiovisual choices are visible, but they are generic, inconsistent, poorly motivated, or only weakly integrated with the content.
- 4 -- Moderate: The presentation is serviceable and partly controlled, but lacks consistent audiovisual purpose or refinement.
- 5 -- Strong: Multiple audiovisual choices are purposeful and reasonably well coordinated. Camera, composition, lighting, or staging clearly strengthen the presentation.
- 6 -- Between the anchors for 5 and 7.
- 7 -- Excellent: The video demonstrates highly controlled, coherent, and expressive visual direction. Its audiovisual choices work together exceptionally well and meaningfully enhance the content.

If the video is corrupted, substantially unavailable, or impossible to evaluate for technical reasons, return `valid=false` and do not assign a score. Keep the evidence and rationale concise and grounded only in observable content.

Return strict JSON with exactly these fields:

```json
{
  "dimension": "cinematic_quality",
  "valid": true,
  "evidence": [
    "<observable evidence 1>",
    "<observable evidence 2>"
  ],
  "score": <your score>,
  "confidence": "high",
  "rationale": "<brief justification grounded in the listed evidence>"
}
```

`valid` must be `true` or `false`. If `valid=false`, `evidence` must describe the technical problem, `score` must be `null`, and `confidence` must be `"low"`. When `valid=true`, `evidence` must contain 1--4 short observable statements, `score` must be an integer from 1 to 7, and `confidence` must be `"high"`, `"medium"`, or `"low"`. Do not output any text outside the JSON object.
\end{lstlisting}

\begin{lstlisting}[style=skillpeprompt,caption={The system prompt for evaluating narrative appeal in four-dimensional video selection.},label={lst:NA}]
You are an expert evaluator of generated videos for text-to-video prompt engineering research.

Evaluate the attached generated video for Narrative Appeal with respect to the original user prompt.

Narrative Appeal measures how effectively the generated video organizes its content into a coherent and engaging temporal experience.

Judge only observable evidence in the generated video. Do not assume that an intended event, state change, transition, pacing effect, or emotion occurred unless it is actually visible. Use only the original user prompt and the generated video. Do not infer or speculate about the generation method, rewritten prompt, skill, model identity, or experimental condition.

First identify concrete temporal evidence, then assign a score. Judge whether events and state changes are understandable, whether transitions make temporal and causal sense, and whether the sequence establishes a readable progression rather than a disconnected collection of moments. Also judge whether pacing, staging, emphasis, or progression produces a meaningful context-appropriate emotional effect such as anticipation, tension, humor, intimacy, surprise, or satisfaction.

Consider:

- clarity of the event progression;
- temporal and causal continuity;
- readability of transitions and state changes;
- pacing and allocation of time;
- buildup, emphasis, and resolution;
- emotional engagement created by the visible progression.

Use this 7-point ordinal scale:

- 1 -- Very poor: The sequence is confusing, fragmented, or emotionally inert. Events do not form an understandable or engaging progression.
- 2 -- Between the anchors for 1 and 3.
- 3 -- Weak: The basic sequence can be understood, but transitions, pacing, audio narration, or emotional progression are weak, abrupt, or poorly organized.
- 4 -- Moderate: The progression is generally understandable, with some effective moments, but engagement or temporal organization remains uneven.
- 5 -- Strong: The video presents a clear and coherent progression with effective pacing and a noticeable degree of emotional engagement.
- 6 -- Between the anchors for 5 and 7.
- 7 -- Excellent: Events, transitions, pacing, audio, and emphasis form an exceptionally coherent and compelling temporal experience with strong, appropriate emotional impact.

If the video is corrupted, substantially unavailable, or impossible to evaluate for technical reasons, return `valid=false` and do not assign a score. Keep the evidence and rationale concise and grounded only in observable content.

Return strict JSON with exactly these fields:

```json
{
  "dimension": "narrative_appeal",
  "valid": true,
  "evidence": [
    "<observable evidence 1>",
    "<observable evidence 2>"
  ],
  "score": <your score>,
  "confidence": "high",
  "rationale": "<brief justification grounded in the listed evidence>"
}
```

`valid` must be `true` or `false`. If `valid=false`, `evidence` must describe the technical problem, `score` must be `null`, and `confidence` must be `"low"`. When `valid=true`, `evidence` must contain 1--4 short observable statements, `score` must be an integer from 1 to 7, and `confidence` must be `"high"`, `"medium"`, or `"low"`. Do not output any text outside the JSON object.
\end{lstlisting}

\begin{lstlisting}[style=skillpeprompt,caption={The system prompt for evaluating creativity in four-dimensional video selection.},label={lst:CR}]
You are an expert evaluator of generated videos for text-to-video prompt engineering research.

Evaluate the attached generated video for Creativity with respect to the original user prompt.

Creativity measures the degree to which the video realizes the user's request through novel, non-obvious, and expressively ambitious visual, audio, or temporal choices while remaining appropriate to the original intent.

Judge only observable evidence in the generated video. Do not assume that an intended idea, viewpoint, transition, visual effect, or expressive choice occurred unless it is actually visible. Use only the original user prompt and the generated video. Do not infer or speculate about the generation method, rewritten prompt, skill, model identity, or experimental condition.

First identify concrete evidence of originality, then assign a score. More shots, stronger camera movement, elaborate lighting, visual effects, unusual content, or greater complexity do not automatically indicate higher Creativity. An unexpected addition is valuable only when it remains compatible with the user's request and contributes meaningfully to the realization.

Consider:

- novelty relative to a straightforward or default realization of the prompt;
- originality in viewpoint, staging, camera language, composition, lighting, transitions, audio, or temporal presentation;
- non-obvious but meaningful expressive choices;
- distinctive integration of multiple creative decisions;
- whether the choices contribute to the expression rather than merely adding complexity.

Do not reward hallucinated subjects or events that contradict the prompt, arbitrary surrealism, random visual artifacts, unnecessary complexity, excessive camera movement, merely using more shots, or deviation for its own sake.

Use this 7-point ordinal scale:

- 1 -- Very poor: The video is a largely literal, default, or generic realization with no clearly identifiable creative treatment.
- 2 -- Between the anchors for 1 and 3.
- 3 -- Weak: The video contains some additional expressive or stylistic choices, but they are mostly conventional, generic, superficial, or weakly integrated.
- 4 -- Moderate: At least one meaningful non-default choice is visible, but the realization is only partly distinctive or consistently developed.
- 5 -- Strong: The video contains clearly non-obvious and appropriate choices that give the realization a distinctive character while preserving the user's intent.
- 6 -- Between the anchors for 5 and 7.
- 7 -- Excellent: The video presents a highly original, distinctive, and expressively ambitious realization, with multiple well-integrated choices that remain coherent and appropriate to the original prompt.

If the video is corrupted, substantially unavailable, or impossible to evaluate for technical reasons, return `valid=false` and do not assign a score. Keep the evidence and rationale concise and grounded only in observable content.

Return strict JSON with exactly these fields:

```json
{
  "dimension": "creativity",
  "valid": true,
  "evidence": [
    "<observable evidence 1>",
    "<observable evidence 2>"
  ],
  "score": <your score>,
  "confidence": "high",
  "rationale": "<brief justification grounded in the listed evidence>"
}
```

`valid` must be `true` or `false`. If `valid=false`, `evidence` must describe the technical problem, `score` must be `null`, and `confidence` must be `"low"`. When `valid=true`, `evidence` must contain 1--4 short observable statements, `score` must be an integer from 1 to 7, and `confidence` must be `"high"`, `"medium"`, or `"low"`. Do not output any text outside the JSON object.
\end{lstlisting}

\subsection{Skill Routing and PE Agent}

At inference time, Gemini 3.1 Pro selects exactly one skill from the chosen library at temperature 0.1. The system prompt for skill selection is shown in Listing \ref{lst:routing}. The same model then performs the rewrite with the selected skill using the system prompt in Listing \ref{lst:pe}.

Note that when generating videos with MiniMax-H3 (except in the length ablation study of Section \ref{sec:abl_length}), we follow the recommended practice to use Gemini 3.1 Pro to adapt the input prompt into its standard input format for better performance. The corresponding system prompt is shown in Listing \ref{lst:h3adaptor}. This adaptation is regardless of whether our proposed approach is used or not, i.e., it applies to both the baseline approaches and our approach in the evaluation.

\begin{lstlisting}[style=skillpeprompt,caption={The system prompt for skill-conditioned rewrite.},label={lst:pe}]
Rewrite <original_prompt> with exactly <selected_skill>. Produce an English, 10-second text-to-video prompt. Preserve subjects, objects, every requested event, event order, and causal relations. Adapt shot rhythm, camera, transitions, scale, and optional sound without replacing content. Specify visible action beats, environment, lighting, framing, and transitions. Do not mention image conditioning. Return JSON: {"pe_prompt": <rewrite>}.
\end{lstlisting}
% Sources: prompts/skill_select.md; prompts/storyeval_skill_pe_rewrite.md;
% prompts/h3_native_prompt_adapter.md; run_storyeval_baseline_eval.py;
% scripts/prepare_v4_0_h3_storyeval_test.py.

\begin{lstlisting}[style=skillpeprompt,caption={The system prompt for prompt adaptation for MiniMax-H3.},label={lst:h3adaptor}]
H3 adapter inputs: <original_prompt>, <skill_conditioned_rewrite>. Convert to the generator's native format while preserving the original request and useful cinematic strategy. Return h3_prompt. Use native fields, in order: integrated_multimodal_description, overall_soundscape, non_diegetic_music. Begin with [Shot 1]; later shots use [Shot N] At 00:SS.mmm, with increasing cut times below 10 seconds. Target 220-420 English words. Complete all events visibly before the end. Validate format and event coverage; repair invalid outputs.
\end{lstlisting}

\section{Additional Details of Evaluation} \label{sec:eval_detail}
Table ~\ref{tab:pe_config} 
summarize the prompt-engineering, video-generation, and evaluation configurations
used in our experiments.

% \textbf{Comparison with Concurrent Work. } WanPE~\citep{zhu2026wanpe} is closely related to our work in its emphasis on cinematic prompt enhancement and shot-level planning. However, its model is not released by the time we submit this work, preventing a faithful direct evaluation using the original trained enhancer. Reproducing its results from published descriptions alone would require recreating its large-scale video-grounded training and semantic-consistency optimization. Furthermore, its reported results on WanPEval use different prompt distributions, video-generation settings, and evaluation protocols, making direct numerical comparisons inappropriate. We therefore avoid treating its published scores as directly comparable and instead include matched-budget optimization baselines in our evaluation.

\section{Additional Study Results}
% \subsection{Sensitivity to Video Generation Seeds}
% \label{app:seed_sensitivity}

% We further evaluate whether the observed gains are sensitive to stochastic video-generation initialization. We randomly sample 50 StoryEval prompts and regenerate videos with MiniMax-H3 using three independently sampled generation seeds, while keeping the prompts, skill libraries, PE outputs, and all other generation settings fixed. The same seed is shared across methods for each prompt to enable paired comparison. In total, this experiment contains 600 generated videos across Direct LLM Rewrite, Seed Skill, SkillPE Top, and SkillPE Top Overall.

\begin{table*}[h]
\centering
\small
\setlength{\tabcolsep}{4pt}
\caption{Prompt-engineering and routing configurations.}
\label{tab:pe_config}
\begin{tabular}{p{2.7cm}p{2.8cm}c p{7.0cm}}
\toprule
\textbf{Method / Stage} &
\textbf{Model / Checkpoint} &
\textbf{Temp.} &
\textbf{Key Configuration} \\
\midrule

Raw Prompt
& --
& --
&  \\

Direct LLM Rewrite
& Gemini 3.1 Pro
& 0.25
& Single-pass rewriting. \\

OPRO & Gemini 3.1 Pro & 0.8 / 0$^{a}$ & OPRO starts from the same 20 normalized seed skills as SkillPE and iterates the skills for four rounds based on the same development set as SkillPE. The computation budget is the same between OPRO and SkillPE (i.e., the same number of  videos are generated). OPRO receives the PF, CQ, NA, and CR scores as feedback during the evolution. \\

VPO
& VPO-5B
& 0
& \texttt{do\_sample=False}\\

Prompt-A-Video
& Prompt\_A\_Video\_CV
& 0
& \texttt{do\_sample=False} \\

Mora: planning / PE
& Gemini 3.1 Pro
& 0.25
& The video generation agent uses MiniMax-H3 or LTX-2.5 for consistency. \\

Mora: first frame
& SDXL Base 1.0
& --
& $1024{\times}576$ resolution, 24 inference steps, guidance scale 7.0; the generated image is used as the frame-0 condition. \\

Seed Skill: routing / PE
& Gemini 3.1 Pro
& 0.1 / 0$^{b}$
& Exactly one skill is selected per prompt.  \\

SkillPE Top Creativity
& Gemini 3.1 Pro
& 0.1 / 0
& Exactly one skill is selected per prompt. \\

SkillPE Top-2 Creativity
& Gemini 3.1 Pro
& 0.1 / 0
& Exactly one skill is selected per prompt. \\

SkillPE Top Overall
& Gemini 3.1 Pro
& 0.1 / 0
& Exactly one skill is selected per prompt. \\

H3-native adaptation
& Gemini 3.1 Pro
& 0
& \\

RAPO: retrieval
& all-MiniLM-L6-v2
& --
& Author-provided retrieval graph, top-3 locations, five neighbors per category, and similarity threshold 0.6. \\

% RAPO: merge / rewrite
% & Mistral-7B-Instruct-v0.3
% & --
% & BF16 inference, iterative modifier merging and direct candidate rewriting. \\

RAPO: refactoring
& llama3\_1\_instruct \_lora\_rewrite
& --
& Author-provided rewrite checkpoint, BF16 inference. \\

RAPO: candidate selection
& Gemini 3.1 Pro
& 0
& Gemini 3.1 Pro replaces the original trained discriminator because the official discriminator model is not public. \\

MiniMax-H3 generation
& MiniMax-H3 Base FL2VA
& --
& $1344{\times}768$, 24 FPS, 10\,s, 50 inference steps, video flow shift 12 and audio flow shift 3. Text-only methods use T2VA, whereas Mora uses FL2VA conditioning. Audio is enabled during generation and delivered as 32-kHz stereo AAC.\\

LTX-2.5 generation
& LTX-2.5 22B Distilled
& --
& Same resolution, duration, frame rate, precision, and $8+3$-step sampler as the standard LTX-2.5 setting, with native prompt enhancement enabled. Audio is enabled during generation and delivered as 32-kHz stereo AAC. \\

Four-dimensional evaluation
& Gemini 3.1 Pro
& 0
& To accommodate the model's input format, the videos are transformed into the following format through FFmpeg before evaluation: Full 10s audiovisual input, 24 FPS with 720-pixel output height, H.264 CRF 23 and 128-kbps stereo AAC. \\

\bottomrule
\end{tabular}

\vspace{2pt}
\begin{minipage}{0.98\textwidth}
\footnotesize
$^{a}$ In OPRO, skill generation and PE use temperatures of 0.8 and 0,
respectively.
$^{b}$ SkillPE routing and rewriting clients explicitly use temperatures of 0.1 and 0,
respectively.
\end{minipage}
\end{table*}

\subsection{Analysis of Benchmark-Native Score Changes}
\label{app:native_metric_analysis}

Our main experiments suggest that in benchmark-native scores, there are non-trivial differences between our SkillPE libraries and the strongest approaches on MiniMax-H3 despite SkillPE's competitive performance on LTX-2.5. Meanwhile, the user study results also reveal a slight decrease in PF on SkillPE Top Creativity (-0.07 points) compared with Seed-skill PE.

% We further analyze the decreases in benchmark-native scores observed after skill evolution. On MiniMax-H3/VBench, the selected libraries reduce the overall native score by 1.30--2.12 points (on a 0--100 scale) relative to Seed Skill. The decrease is larger on the semantic component ($-2.97$ to $-6.47$) than on the quality component ($-0.79$ to $-1.32$). The largest contributors include scene consistency, multiple-object and spatial-relation alignment, together with smaller degradations in imaging quality, temporal flickering, and background consistency. At the same time, some dimensions improve. For example, Top Overall increases dynamic degree by 8.33 points.

% A similar trade-off appears on StoryEval, where the selected libraries reduce the native event-completion score by 1.92--2.86 points relative to Seed Skill. The decrease is more pronounced on the hard subset ($-3.39$ to $-4.83$), indicating that stronger cinematic treatment can occasionally interfere with faithful realization of requested events.

Our manual inspection of the failed cases suggests one possible explanation: more expressive cinematic
realizations introduce additional camera movements, viewpoint changes, and shot transitions, which may not be well aligned with the benchmark-native metrics naturally. Since some VBench
consistency metrics rely on cross-frame feature similarity (e.g., \texttt{Subject Consistency}),
such changes may affect their scores even when they serve
an intentional cinematic purpose. Nevertheless, more
complex shot structures may also reduce explicit evidence of event completion (e.g., the realization of some event may be instant or implicitly inferred based on relevant scenes, but no clear and persistent visual scenes are present). We therefore envision two future directions based on our insights. First, we envision stronger video generation models to preserve better evidence-aware user fidelity and visual consistency while enhancing cinematic appeal with SkillPE. Second, future benchmarks can provide better video quality assessment for cinematic videos with implicit events, multi-shot design, or complicated visual effect.

\subsection{Length Distribution of PE Result}
To complement the ablation study on prompt length (Section \ref{sec:abl_length}), Table \ref{tab:prompt_length} shows the length distribution of the PE results for each baseline in our study. 
% Please note that after the shared backbone-specific adaptation, the final prompt lengths are comparable across methods with only minor differences.

\begin{table}[t]
\centering
\small
\setlength{\tabcolsep}{5pt}
\caption{Prompt length (mean $\pm$ standard deviation, in whitespace-delimited English words) distribution before backbone-specific prompt adaptation.}
\label{tab:prompt_length}
\begin{tabular}{lcc}
\toprule
\textbf{Method} & \textbf{StoryEval} & \textbf{VBench} \\
\midrule
Raw Prompt          & $16.1\pm4.5$   & $7.6\pm5.9$ \\
Direct LLM Rewrite  & $43.2\pm12.6$  & $52.7\pm19.0$ \\
OPRO                 & $191.6\pm89.9$  & $183.1\pm100.3$ \\
VPO                 & $105.8\pm8.5$  & $101.9\pm6.8$ \\
Prompt-A-Video      & $99.1\pm12.0$  & $98.1\pm11.3$ \\
RAPO                & $43.4\pm11.4$  & $40.6\pm14.2$ \\
Mora                & $29.9\pm15.4$  & $28.0\pm15.3$ \\
Seed Skill & $113.0\pm50.2$ & $113.5\pm52.2$ \\
\midrule
\multicolumn{3}{l}{\textit{Libraries selected with MiniMax-H3}} \\
SkillPE Top Creativity        & $171.0\pm67.2$ & $191.9\pm94.6$ \\
SkillPE Top-2 Creativity       & $216.4\pm76.2$ & $186.2\pm88.0$ \\
SkillPE Top Overall     & $178.6\pm63.3$ & $178.5\pm73.8$ \\
\midrule
\multicolumn{3}{l}{\textit{Libraries selected with LTX-2.5}} \\
SkillPE Top Creativity         & $162.4\pm76.6$ & $151.0\pm73.0$ \\
SkillPE Top-2 Creativity     & $182.3\pm85.9$ & $182.2\pm92.3$ \\
SkillPE Top Overall     & $166.1\pm90.8$ & $152.0\pm86.2$ \\
\bottomrule
\end{tabular}
\end{table}

\begin{lstlisting}[style=skillpeprompt,caption={The preamble of the system prompt for reference-free R\&D-based skill refinement.},label={lst:RF}]
You are an expert in prompt engineering for video generation. Based on your own knowledge, your task is to add practical usage guidance to an existing seed skill.

When evolving the raw seed, think for each of the dimensions about what should be added and when it should be strengthened based on your own knowledge. Likewise, think about what should not be added and when the skill should not be selected. Return valid JSON with the following fields: [The remaining part is the same as Listing 4]
\end{lstlisting}

\subsection{Ablation of Movie Reference Guidance}
\label{sec:reference_ablation}

To examine whether movie references provide additional benefits
beyond LLM-based skill evolution and assessed selection, we
conduct a reference-free ablation. Specifically, we remove the
retrieved movie references during skill evolution, allowing
the LLM to refine and mutate the normalized seed skills
using its internal knowledge alone. In particular, we change the preambles of the system prompts for R\&D-based skill refinement following Listing \ref{lst:RF}. For the system prompt of divergent-based skill evolution, we replace  ``\texttt{Ground at least one major change in a divergent reference}'' with ``\texttt{Ground the changes based on your own knowledge}''.
We retain the same
candidate budget, development set, and video-based
skill selection procedure. The resulting libraries are
evaluated under the same setting as SkillPE.

\begin{table}[t]
    \centering
    \caption{Ablation of movie-reference guidance on
    StoryEval with MiniMax-H3. ``Reference'' indicates whether
    movie references are used during skill evolution.}
    \label{tab:reference_ablation}
    \small
    \setlength{\tabcolsep}{3pt}
    \begin{tabular}{llccccc}
        \toprule
        Selection & Reference & PF & CQ & NA & CR & Overall \\
        \midrule
        \multirow{2}{*}{Top Creativity}
          & No  & \textbf{6.41} & 5.72 & 5.71 & 4.38 & 5.56 \\
          & Yes & 6.29 & \textbf{6.51} & \textbf{5.87} &
          \textbf{5.13} & \textbf{5.95} \\
        \midrule
        \multirow{2}{*}{Top-2 Creativity}
          & No  & 6.39 & 5.76 & 5.60 & 4.45 & 5.55 \\
          & Yes & \textbf{6.40} & \textbf{6.60} &
          \textbf{5.82} & \textbf{4.89} & \textbf{5.93} \\
        \midrule
        \multirow{2}{*}{Top-Overall}
          & No  & \textbf{6.37} & 5.82 & 5.67 & 4.30 & 5.54 \\
          & Yes & 6.38 & \textbf{6.64} & \textbf{5.98} &
          \textbf{5.17} & \textbf{6.04} \\
        \bottomrule
    \end{tabular}
\end{table}

As shown in Table~\ref{tab:reference_ablation}, reference-guided evolution consistently outperforms its reference-free counterpart across all three library selection strategies, improving the Overall score by 0.38--0.50 points. The improvements are primarily driven by substantial gains in Cinematic Quality (+0.79--0.84) and Creativity (+0.44--0.87), with smaller gains in Narrative Appeal (+0.16--0.31). Meanwhile, Prompt Fidelity remains roughly stable, with changes ranging from -0.12 to +0.01, reflecting the trade-off between creative expression and semantic preservation. These results demonstrate the importance of movie-reference guidance in skill evolution, suggesting that real cinematic examples provide valuable inspiration for developing more creative and narratively engaging skills beyond reference-free LLM-based refinement and mutation.

\subsection{Cross-backbone Generalization of Skills}
Given that the backbone model used for skill selection during evolution is the same as the one used for evaluation in the main study, we further assess whether the skills are reusable enough to generalize to other backbones not used for selection. Specifically, we assess the evolved libraries (Top Creativity, Top-2 Creativity, and Top Overall) selected by MiniMax-H3 on LTX-2.5, and assess the ones selected by LTX-2.5 on MiniMax-H3.

Table \ref{tab:backbone-selection-eval} shows the results. Across both benchmarks, all evolved skill libraries outperform the seed skills on backbones unused for selection. This demonstrates that SkillPE transfers with positive gain over seed.  

\begin{table}[htbp]
\centering
\small
\caption{Evaluation results of generalizing evolved skills to backbones not used for selection. ``LTX Top Creativity $\rightarrow$ H3'' means evaluating the Top Creativity skill library selected by LTX on MiniMax-H3. All scores are Overall scores.}
\label{tab:backbone-selection-eval}
\begin{tabular}{lcc}
\toprule
Skill \& Backbone & StoryEval ($\Delta$) & VBench ($\Delta$) \\
\midrule
Seed Skill $\rightarrow$ H3        & 5.11  & 5.09  \\
LTX Top Creativity $\rightarrow$ H3       & 5.87 (+0.76) & 5.65 (+0.56) \\
LTX Top-2 Creativity $\rightarrow$ H3     & 5.85 (+0.74) & 5.63 (+0.54) \\
LTX Top Overall $\rightarrow$ H3          & 5.89 (+0.78) & 5.59 (+0.50) \\
\midrule
Seed Skill $\rightarrow$ LTX       & 4.46  & 4.78  \\
H3 Top Creativity $\rightarrow$ LTX       & 4.91 (+0.45) & 5.54 (+0.76) \\
H3 Top-2 Creativity $\rightarrow$ LTX     & 5.16 (+0.70) & 5.47 (+0.69) \\
H3 Top Overall $\rightarrow$ LTX          & 4.88 (+0.42) & 5.76 (+0.98) \\
\bottomrule
\end{tabular}
\end{table}

\subsection{Robustness of Human Evaluation}
\label{app:human_robustness}

To additionally account for uncertainty arising from both prompt and annotator sampling, we perform 10,000 two-way bootstrap replicates by independently resampling the 40 prompts and ten annotators with replacement while preserving the matched method structure. As shown in Table~\ref{tab:human_bootstrap}, all three final SkillPE libraries retain positive improvements over Seed-skill PE, with the 95\% confidence intervals remaining above zero.

\begin{table}[t]
\centering
\small
\caption{Robustness of the human evaluation under two-way bootstrap resampling over prompts and annotators. Intervals are pointwise 95\% percentile confidence intervals.}
\label{tab:human_bootstrap}
\begin{tabular}{lc}
\toprule
\textbf{Comparison} & \textbf{$\Delta$ 4D Avg. [95\% CI]} \\
\midrule
SkillPE Top Creativity$-$ Seed     & $+0.560$ [$0.047$, $1.078$] \\
SkillPE Top-2  Creativity$-$ Seed   & $+0.568$ [$0.084$, $1.038$] \\
SkillPE Top Overall $-$ Seed & $+0.805$ [$0.341$, $1.288$] \\
\bottomrule
\end{tabular}
\end{table}

\section{Case Study} \label{sec:case_study}
% We present three complementary case studies to illustrate (i) how SkillPE  progressively enriches prompts along the full pipeline,  (ii) how the adaptive skill discovery shapes the eventual  prompts and videos, and (iii) the effectiveness and trade-offs of SkillPE's evolution.

We provide additional qualitative analysis complementing the discussion in Section \ref{sec:case_study_main}. 

\begin{figure*}[t]
  \centering
  % \scalebox{0.86}{
  \includegraphics[width=\linewidth]{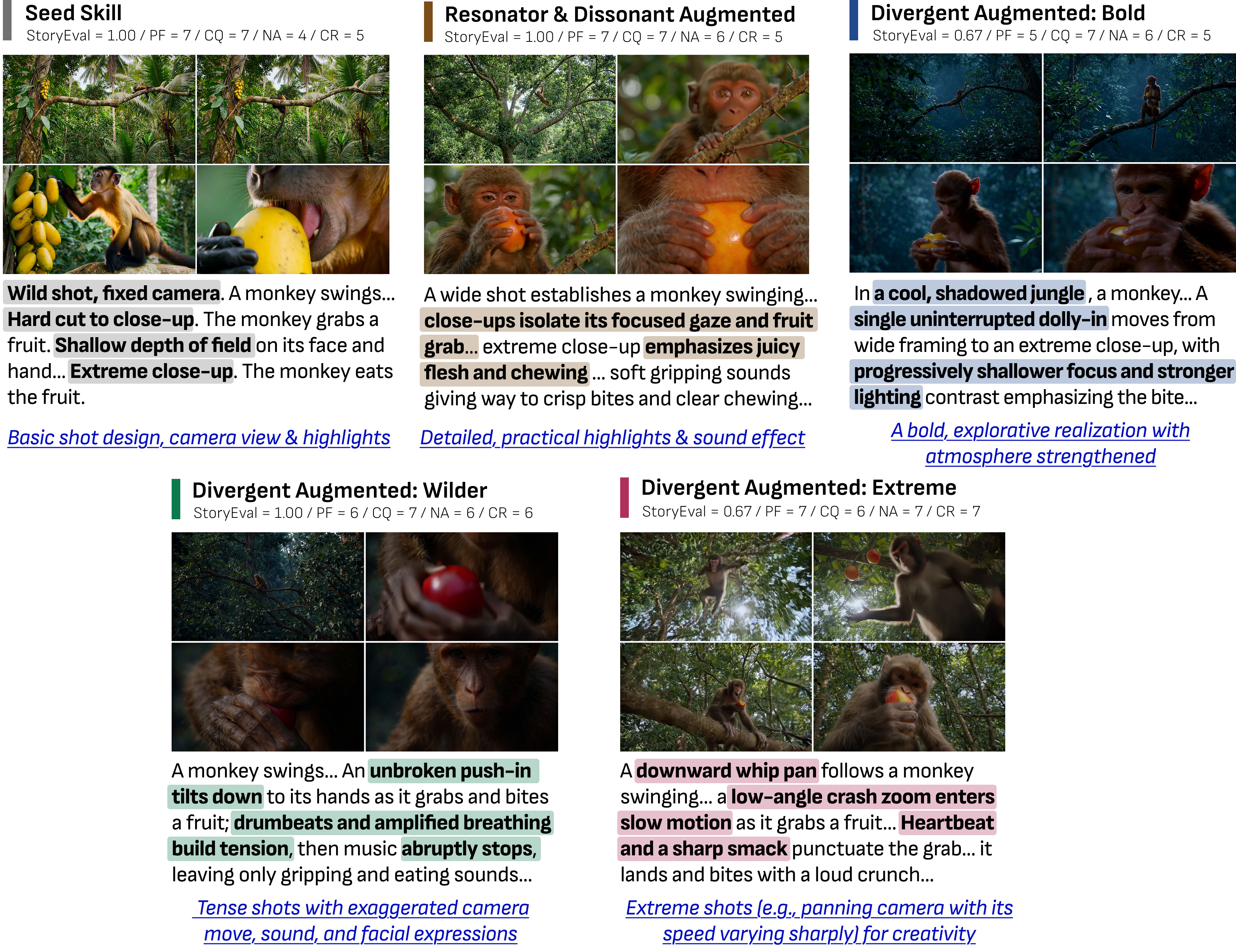}
  \caption{Examples that reveal the effect of different candidate skills in the adaptive skill discovery.}
  \label{fig:ablation_case}
\end{figure*}

% \textbf{Progressive prompt enrichment.} Figure \ref{fig:good_case} shows two examples from StoryEval with MiniMax-H3 as the backbone, comparing raw prompts, seed-skill PE, and final SkillPE. In the car-to-robot scenario, seed-skill PE introduces steady frontal tracking, centered composition, and explicit action ordering, with metallic sounds accompanying the transformation. Final SkillPE adds continuous orbiting camerawork, which contrasts significantly  the steady frontal tracking in the previous results and contributes to a creative visual narration. It further incorporates
% detailed mechanical unfolding and waving while walking, integrating the requested events into a more dynamic sequence. These refinements preserve prompt fidelity while improving cinematic quality (6 to 7), narrative appeal (5 to 6), and creativity (2 to 6).

% In the book-and-fish scenario, seed-skill PE builds anticipation through cuts from a wide establishing shot to close-ups of the man's face and the book, supported by a building heartbeat and rising musical chord. Final SkillPE strengthens the dramatic contrast: tight close-ups in darkness establish tension before a wide shot reveals blazing light and swirling fish. The prompt further specifies a rising hum that cuts to silence before an explosive boom, reinforcing the intended transition from suspense to wonder. Cinematic quality remains at 7, while narrative appeal and creativity both increase from 6 to 7, illustrating how further enrichment strengthens dramatic impact beyond an already coherent cinematic presentation.

\textbf{Adaptive Skill Discovery.} Figure~\ref{fig:ablation_case} illustrates how different evolution strategies transform the prompt ``\textit{A monkey swings from one tree to another, grabs a fruit, and then eats it.}'' The Seed Skill establishes basic shot design, camera viewpoints, and visual highlights. Resonator and Dissonant (R\&D) augmentation enriches these foundations with practical action details, focused close-ups, and synchronized sound effects, improving Narrative Appeal from 4 to 6. Divergent evolution explores increasingly expressive cinematic realizations: Bold introduces a cool, shadowed atmosphere through a continuous dolly-in, progressively shallower focus, and stronger lighting contrast. Wilder builds tension through exaggerated camera movements, expressive close-ups, amplified breathing, and dramatic sound transitions. Extreme employs a downward whip pan, a low-angle crash zoom, slow motion, and impactful sound effects, creating the most dynamic realization and achieving the highest Narrative Appeal (7) and Creativity (7). These examples illustrate how reference-guided evolution expands cinematic possibilities beyond basic shot design.

% \textbf{Analysis of Evolution Effectiveness and Trade-Offs.} Figure~\ref{fig:ablation_case} also reveals the complementary effects and trade-offs of different evolution strategies. R\&D improves narrative presentation while preserving Prompt Fidelity and Cinematic Quality. Among divergent variants, Creativity increases from 5 (Bold) to 6 (Wilder) and 7 (Extreme), but other metrics do not improve uniformly. Bold emphasizes atmosphere at the expense of Prompt Fidelity (5), while Wilder achieves stronger Creativity (6). Extreme further enhances Narrative Appeal and Creativity while maintaining Prompt Fidelity (7), but slightly reduces Cinematic Quality (6) and the official StoryEval score (0.67). These results highlight the importance of exploring diverse cinematic strategies and employing video-based assessment to balance creative expressiveness with faithful narrative realization.

\end{document}